\documentclass[11pt, a4paper,copyright, goog]{google}

\usepackage[authoryear, sort&compress, round]{natbib}
\usepackage{xspace}
\usepackage{wrapfig}
\usepackage{titletoc}
\usepackage{multirow}
\usepackage{lipsum}
\newcommand{\method}{\textsc{EnvHarness}\xspace}
\newcommand{\designer}{\textsc{EnvRigger}\xspace}
\newcommand{\sd}[1]{\,$_{\textcolor{gray}{#1}}$}

\definecolor{hlcolor}{HTML}{8B1A1A}
\newcommand{\evocolor}[1]{{\color{hlcolor}#1}}
\usepackage[most]{tcolorbox}
\definecolor{tealline}{HTML}{0E9488}   
\definecolor{tealfill}{HTML}{F0F9F7}   
\definecolor{amberline}{HTML}{B45309}  
\definecolor{amberfill}{HTML}{FDF6EC}  
\definecolor{slate}{HTML}{64748B}      
\definecolor{ink}{HTML}{1E293B}        
\newtcblisting{componentbox}[1]{
  enhanced, breakable,
  listing only,
  colback=amberfill,
  frame hidden, boxrule=0pt, arc=2pt,
  borderline west={2.5pt}{0pt}{amberline},
  left=9pt, right=9pt, top=6pt, bottom=6pt,
  before skip=2pt, after skip=10pt,
  title={\small\bfseries\textcolor{ink}{#1}},
  colbacktitle=amberfill, toptitle=5pt, bottomtitle=1pt,
  titlerule=0pt,
  listing options={
    language=Python,
    basicstyle=\ttfamily\footnotesize,
    breaklines=true,
    columns=fullflexible,
    keepspaces=true,
    showstringspaces=false,
    commentstyle=\itshape\color{slate},
  }
}
\definecolor{pmtink}{HTML}{1E293B}
\definecolor{pmtkey}{HTML}{0E9488}
\definecolor{pmtdim}{HTML}{94A3B8}
\definecolor{pmtline}{HTML}{CBD5E1}
\definecolor{pmtback}{HTML}{FBFCFD}

\newtcolorbox{promptbox}{
  enhanced, breakable,
  colback=slatefill,
  frame hidden, boxrule=0pt, arc=2pt,
  borderline west={2.5pt}{0pt}{slateline},
  left=9pt, right=9pt, top=7pt, bottom=7pt,
  before skip=8pt, after skip=2pt,
  title={\small\bfseries\textcolor{ink}{System prompt of the \method designer agent}},
  colbacktitle=slatefill, toptitle=6pt, bottomtitle=1pt,
  titlerule=0pt,
}

\lstdefinestyle{prompt}{
  basicstyle=\ttfamily\scriptsize\color{ink},
  emph={rules_code,in_env_actions,filter_action,modify_transition,
        filter_observation,_Rules,Rules,env_state,ACCEPT,REFINE,REJECT,
        BASELINE,PITFALL,OBJECTIVE},
  emphstyle=\color{slateline}\bfseries,
  breaklines=true, breakindent=0pt, breakatwhitespace=false,
  columns=flexible, keepspaces=true, showstringspaces=false,
  upquote=true, aboveskip=0pt, belowskip=0pt,
}
\definecolor{slateline}{HTML}{64748B}
\definecolor{slatefill}{HTML}{F4F6F9}
\definecolor{hdr}{HTML}{EFF3F6}
\newtcolorbox{weaknessbox}{
  enhanced, breakable,
  colback=slatefill,
  frame hidden, boxrule=0pt, arc=2pt,
  borderline west={2.5pt}{0pt}{slateline},
  left=9pt, right=9pt, top=7pt, bottom=7pt,
  before skip=8pt, after skip=2pt,
  fontupper=\small,
  title={\small\bfseries\textcolor{ink}{Specified weakness}},
  colbacktitle=slatefill, toptitle=6pt, bottomtitle=1pt,
  titlerule=0pt,
}

\newcommand{\skillorigin}[1]{\par\smallskip\noindent{\footnotesize\itshape\textcolor{slate}{Origin: #1}}}
\tcbset{pairtop/.style={after skip=0pt, sharp corners=south}}
\newtcolorbox{skillbox}[2][]{
  enhanced, breakable,
  colback=tealfill,
  frame hidden, boxrule=0pt, arc=2pt,
  borderline west={2.5pt}{0pt}{tealline},
  left=9pt, right=9pt, top=7pt, bottom=7pt,
  before skip=8pt, after skip=8pt,
  fontupper=\small,
  colbacktitle=tealfill, toptitle=6pt, bottomtitle=1pt,
  titlerule=0pt,
  title={\small\bfseries\textcolor{ink}{#2}},
  #1
}
\newcommand{\skillfield}[1]{\textit{\textcolor{slate}{#1:}}~}

\uselogo{} 

\title{\method: Awakening Static Worlds for Agent Learning}

\correspondingauthor{chengsong@wustl.edu, \{zifengw, chenyulee\}@google.com}

\author[1*]{Chengsong Huang}
\author[2]{Zifeng Wang}
\author[2]{Rujun Han}
\author[2]{Jun Yan}
\author[2]{Yanfei Chen}
\author[2]{Zoey CuiZhu}
\author[2]{Ke Jiang}
\author[4]{Peng Xia}
\author[2]{Han Yu}
\author[2]{Yufan Zhuang}
\author[2]{Yifei Ming}
\author[3]{Jiaqi Pan}
\author[2]{Bhavana Dalvi Mishra}
\author[1]{Jiaxin Huang}
\author[2]{Burak Gokturk}
\author[2]{Tomas Pfister}
\author[2]{Chen-Yu Lee}
\affil[1]{Washington University in St. Louis}
\affil[2]{Google Cloud AI Research}
\affil[3]{Google Cloud}
\affil[4]{University of North Carolina at Chapel Hill}

\begin{abstract}
LLM agents learn by interacting with environments, yet these environments are hand-built and static: blind to an agent's weaknesses, and quickly left behind as it improves. While recent environment generation methods attempt to address this, they require domain-specific pipelines, rely on expensive or unreliable verifiers, and still produce static environments. To alleviate the engineering burden of rebuilding environments from scratch, we propose Environment Harness (\method{}), a programmable layer of plug-in components that wraps a static environment to reshape its behavior without modifying the underlying logic. Operating through standard interfaces, \method{} applies across diverse domains while ensuring every reshaped environment retains its original verifier. To automate this process, we introduce  \designer{}, which treats the target policy as a black box, observing its execution trajectories to synthesize \method{} components targeting diagnosed flaws, and validating them via fresh rollouts. Across five benchmarks in four domains, \method{} outperforms both original environments and domain-specific environment generation pipelines, achieving up to a 9.0-point improvement on held-out instances with 9.8\% fewer execution steps. Furthermore, \method{} provides a superior optimization signal for reinforcement learning, enabling continuous, targeted co-evolution of the policy and its environment.

\vspace{4pt}
{\centering
\raisebox{-0.16\height}{\includegraphics[height=1.05em]{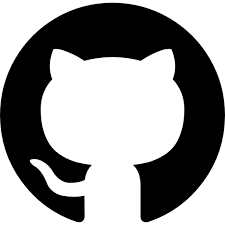}}\hspace{0.32em}\href{https://github.com/google-research/envharness}{\texttt{github.com/google-research/envharness}}%
\hspace{1.15em}\textcolor{gray!45}{\rule[0.02em]{0.6pt}{0.9em}}\hspace{1.15em}%
\raisebox{-0.16\height}{\includegraphics[height=1.05em]{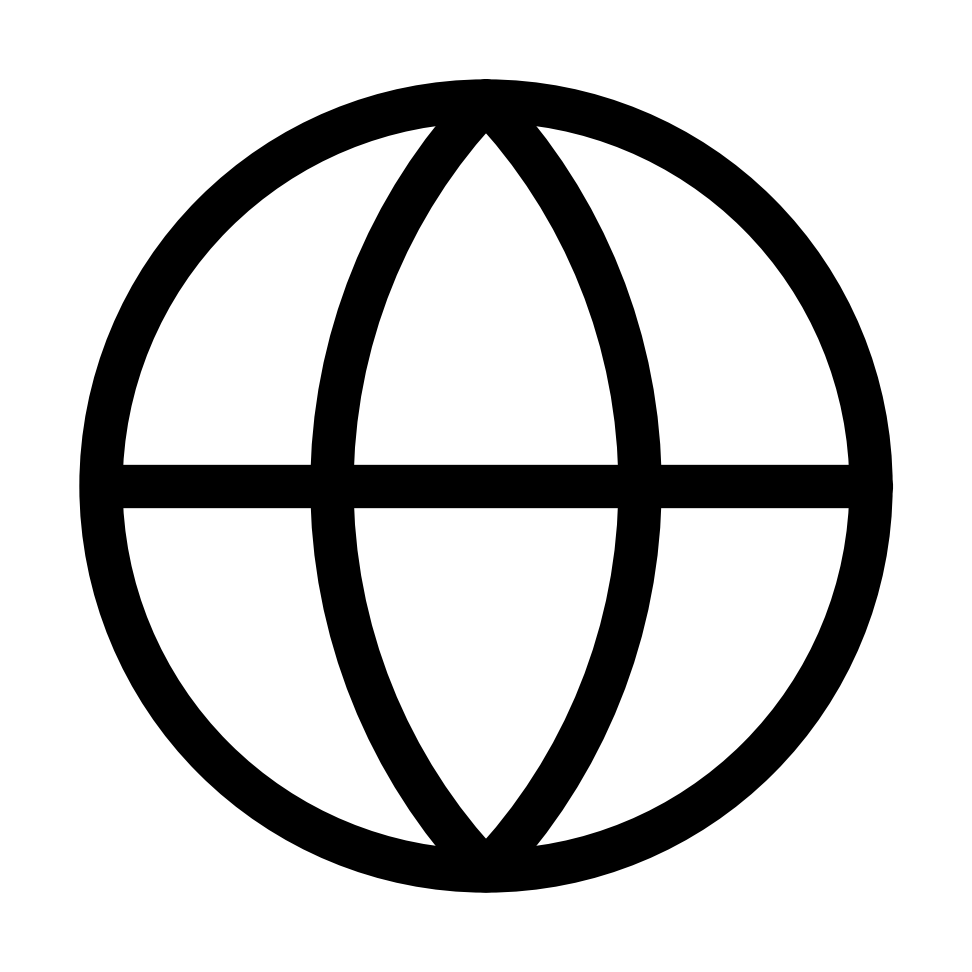}}\hspace{0.32em}\href{https://www.envharness.com}{\texttt{www.envharness.com}}%
\par}

\end{abstract}

\begin{document}

\maketitle

\label{sec:intro}
\begin{figure*}[ht]
  \centering
  \includegraphics[width=0.96\textwidth]{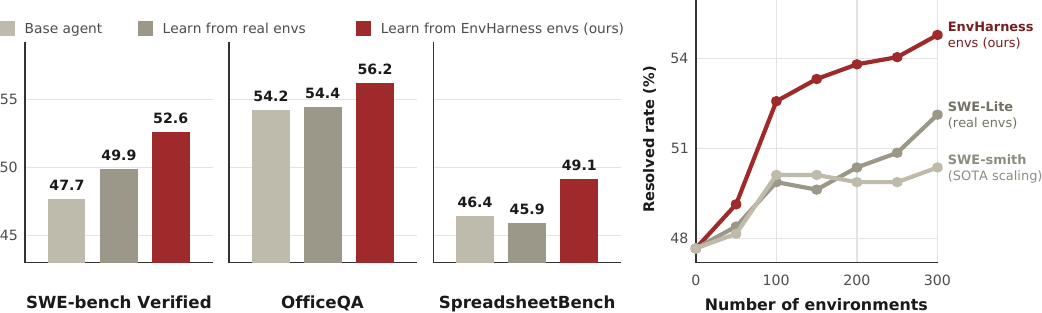}
\caption{\textbf{Overall performance.} \textit{Left:} 
Agents learning from \method environments consistently outperform those learning from the original environments across software engineering and office automation benchmarks, including SWE-bench Verified, OfficeQA, and SpreadsheetBench.
\textit{Right:} On SWE-bench Verified, under an identical environment budget,
\textsc{EnvHarness} keeps improving as environments scale, while real and
generated environments flatten out.}
  \label{fig:new1}
\end{figure*}

\section{Introduction}
\label{sec:intro}
\begin{figure*}[ht]
  \centering
  \includegraphics[width=0.85\textwidth]{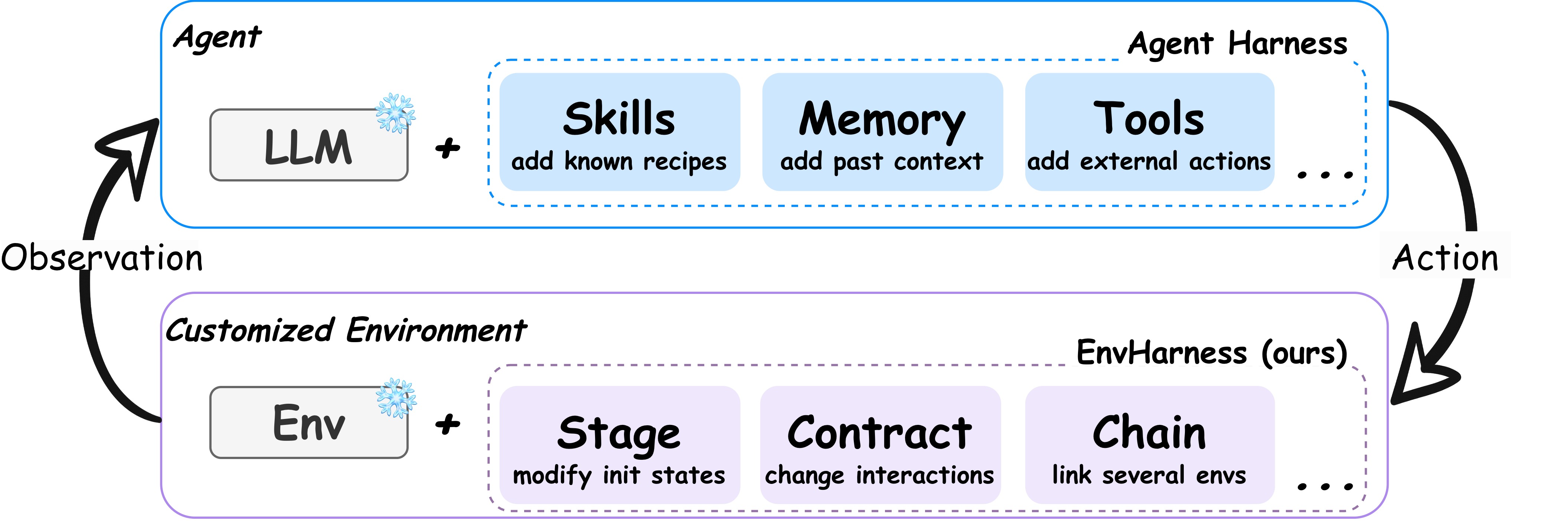}
\caption{
  While an agent harness transforms a frozen LLM into a capable agent via plug-in components (e.g., skills, memory, tools) without altering model weights, \method{} applies this same principle to the other side of the interaction. It customizes a frozen environment with plug-in components while leaving original environment unchanged.
}
  \label{fig:teaser}
\end{figure*}
As LLMs are deployed as autonomous agents, the source of learning shifts from curated text data to interactive environments. Whether navigating web pages~\citep{gur2024real}, resolving issues in a codebase~\citep{yang2024swe}, or controlling an embodied platform~\citep{wang2023voyager}, agents rely on their respective environments to acquire learning signals~\citep{yao2022react}. These environments act as interactive counterparts that present specific tasks, manage changing states, respond to actions, and evaluate success~\citep{li2026agentic}. Unfortunately, building them requires substantial human effort to hardcode the interaction logic and verifiers~\citep{zhang2025swebenchgoeslive, merrill2026terminal}. Consequently, the resulting environments remain rigidly static, behaving identically regardless of which agent interacts with them or how much that agent has improved~\citep{hu2026seal}. This rigidity limits agent learning in two fundamental ways: it fails to provide a targeted signal that addresses a particular agent's weaknesses~\citep{dennis2020emergent, jiang2021prioritized}, and it has nothing more to teach once the agent learns to solve the existing tasks~\citep{wang2019paired, beukman2024refining}.

Since manually building environments is expensive, a growing line of work has turned to automated environment generation~\citep{zala2024envgen, guo2025genenv, song2026envscaler}. Despite its scalability, this approach suffers from two major limitations. First, generation pipelines are inherently domain-specific. Existing systems generate environments for web navigation~\citep{trabucco2025insta}, programming~\citep{pan2024training}, or tool use~\citep{lee2026environment}, but a pipeline built for one setting cannot transfer to the others. 
Second, ensuring correctness is both costly and unreliable. Because these environments and verifiers are generated by LLMs, practitioners must over-generate and heavily filter them~\citep{wang2026agent, yang2026swe}, which still cannot fully guarantee their correctness.

Rather than creating new environments to obtain learning signals, we propose Environment Harness \textbf{(\method)}, a programmable layer that transforms an existing, static environment into a dynamically customized one without modifying the environment itself. We make an analogy between \method and agent harness~\citep{anthropic2025effective, anthropic2026harness, lopopolo2026harness} in Figure~\ref{fig:teaser}: An agent harness provides LLMs with external memory, tools, and skills to handle complex tasks beyond basic text generation. \method extends this concept to environments, equipping a static environment with modular, plug-in components. A \texttt{Stage} sets the starting point of an episode, a \texttt{Contract} controls the allowed actions and observations, and a \texttt{Chain} connects multiple base environments to form an extended episode. 
The agent continues to use the standard interface, but \method mediates this interaction. This enables a single environment to fulfill needs it was never built for: isolating a specific skill, extending a task's horizon, or calibrating difficulty so the agent struggles but ultimately succeeds. Crucially, operating strictly at the interface level makes \method domain-agnostic, while allowing every new environment to safely inherit the trusted, human-built verifier of its original environment.

While \method{} provides a universally applicable framework, its specific configuration must be tailored to each target policy and task. To automate this customization process, we introduce \designer{}. By treating the policy strictly as a black box, \designer{} observes both successful and failed execution trajectories within the base environment to diagnose specific behavioral vulnerabilities. Guided by these findings, it synthesizes candidate \method{} components and wraps the current environment to evaluate them. \designer{} then runs fresh policy rollouts, judging acceptance solely on whether the candidate environment effectively cultivates the missing capabilities while remaining solvable. Components failing this evaluation undergo iterative revision until they succeed. Ultimately, this workflow realizes fully automated, task-policy-conditioned environment customization.

We conduct experiments on challenging benchmarks spanning embodied tasks (ALFWorld~\citep{shridhar2020alfworld}), web browsing (WebArena~\citep{zhou2024webarena}), software engineering (SWE-bench Verified~\citep{jimenez2024swebench}), and office work (OfficeQA~\citep{officeqa}, SpreadsheetBench~\citep{ma2024spreadsheetbench}). Specifically, we evaluate \method across two representative learning paradigms: skill-based learning (SL) and reinforcement learning (RL). In SL settings, agents trained with \method environments outperform those trained on original environments, achieving up to 9.0 points of improvement on held-out tasks (Table~\ref{tab:main_results}) while using 9.8\% fewer interaction steps (Table~\ref{tab:main_results_swe}). In RL settings, \method-customized environments similarly yield significantly stronger policies, with up to 6.5 points of improvement (Table~\ref{tab:rl_results}). Finally, repeatedly executing the \designer loop enables continuous co-evolution between the agent and its environment, unlocking compounding performance gains that scale effectively with the number of customized tasks.

Our contributions are threefold: (1) We propose \method, a programmable layer that customizes a static environment into a controllable one through its own \texttt{reset}/\texttt{step} interface, instantiated as three types of plug-in components that reshape environment initial states, agent-environment interaction interfaces, and composite tasks from different environments, all while the original environment's tasks and verifiers stay unchanged. (2) We introduce \designer{} to automate task-policy-conditioned environment customization. By diagnosing policy flaws from rollouts and iteratively revising candidate components until fresh rollouts confirm their success, \designer{} ensures that every new environment targets the specific weaknesses of the policy.
(3) Across five benchmarks in four domains, \method improves both effectiveness (up to 9.0 points on held-out tasks) and efficiency (9.8\% fewer steps), strengthens policies under reinforcement learning, and scales agent performance where both human-built and generated environments flatten out.

\section{\method}

\subsection{The \method Paradigm}
\label{sec:envharness_paradigm}

\begin{table}[ht]
\centering
\caption{Analogy between Agent Harness and \method. Both scale capabilities through external layers rather than changing the core system.}
\label{tab:harness_comparison}
\resizebox{\linewidth}{!}{
\begin{tabular}{lll}
\toprule
 & \textbf{Agent Harness} & \textbf{\method (Ours)} \\
\midrule
\textbf{Base System}      & Frozen LLM & Static environment \\
\textbf{Designed to Solve}   & Lack of action, memory, or loops & Hardcoded interaction logic \\
\textbf{Harness Layer}    & Capabilities (tools, memory) & Customization (states, rules, observations) \\
\textbf{Unified Output}   & Autonomous agent & Customized environment \\
\bottomrule
\end{tabular}
}
\end{table}

An \emph{agent harness}~\citep{anthropic2025effective, anthropic2026harness, lopopolo2026harness} is the software layer (execution loops, tool registries, and context management)~\citep{meng2026agentharness} that wraps a LLM to form an autonomous agent ($\text{Agent} = \text{Model} + \text{Harness}$). It adds new capabilities without changing the model weights. We apply the same idea to the other side of the agent-environment loop. We define \textbf{\method} as a programmable layer that wraps an existing, static environment and turns it into a customizable one ($\text{Customized Env} = \text{Static Env} + \text{\method}$). \method achieves this customization by modifying the information flow through the standard interface, leaving the underlying environment completely untouched. Analogous to the tools and memory of an agent harness, \method is assembled from modular plug-in components that tailor the environment to specific training needs like isolating a specific skill, extending a task's horizon, or adjusting task difficulty.

\paragraph{Formal Definition of \method.}
We model an environment as a tuple $E = (\mathcal{S}, \mathcal{A}, \mathcal{O}, T, R, s_0)$, where $\mathcal{S}$ is the state space, $\mathcal{A}$ the action space, $\mathcal{O}$ the observation space, $T: \mathcal{S}\times\mathcal{A}\rightarrow \mathcal{S}$ the transition function that maps a state and an action to the next state, $R$ the reward induced by the verifier, and $s_0$ the initial state. An \method{} component is an environment-agnostic transformation $w$:

\begin{equation}
E' = w(E), \qquad E' = (\mathcal{S}', \mathcal{A}', \mathcal{O}', T', R', s_0').
\label{eq:wrapper}
\end{equation}

$w$ reshapes the environment strictly at the interface level, without modifying its underlying simulator backend or implementation details. It customizes the initial state ($s_0'$), filters exposed spaces ($\mathcal{A}', \mathcal{O}'$), and updates transition mechanics ($T'$). Because all interventions remain external, the ground-truth evaluation logic is preserved, ensuring the original verifier still can score the episode. 

\subsection{Three \method Components}
\label{sec:components}

\begin{figure}[h]
    \centering
    \includegraphics[width=0.9\linewidth]{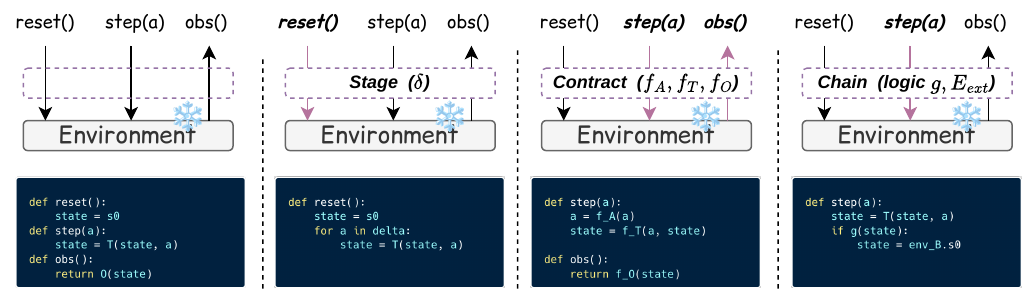}
    \caption{\textbf{Overview of \method components wrapping the standard environment interface.} The underlying base environment (native state transitions and original task verifier) remains completely frozen. From left to right: the base environment, followed by three \method components—\texttt{Stage}, \texttt{Contract}, and \texttt{Chain}. Highlighted arrows and headers indicate overridden interface methods, with code blocks showing how each wrapper modifies state initialization, transition dynamics, or observation handling without altering the base environment.}

    \label{fig:components}
\end{figure}

The mapping $w$ of Eq.~\eqref{eq:wrapper} defines a general interface, and any transformation that follows this interface is a valid \method{} component. In this work, we introduce three concrete types of components, chosen to cover three fundamental modes of environment customization, and we expect more to follow.
Each type is specified by its own parameters, overriding standard environment interface methods such as \texttt{reset} or \texttt{step}. Figure~\ref{fig:components} compares the three component types with the original interface, highlighting the specific method each component overrides. We detail the software implementation of this interface protocol and class architecture in Appendix~\ref{sec:interface}. Throughout this subsection, we illustrate the components on one ALFWorld task, \textbf{\emph{``put a clean mug on the desk''}}, where the default instance leaves the mug in the open and ends immediately upon placement.

\paragraph{\texttt{Stage}: changing the initial state.}
A \texttt{Stage}, $w_{\mathrm{stage},\delta}$, is specified by a sequence of state-manipulation actions $\delta = (a_1, \ldots, a_k)$.
These actions are applied to the initial state $s_0$ produced by \texttt{reset()}:
\begin{equation}
E' = w_{\mathrm{stage}, \delta}(E) = (\mathcal{S}, \mathcal{A}, \mathcal{O}, T, R, \evocolor{s_0'}), \quad \text{where } \evocolor{s_0'} = T\left( \cdots T\left(T(s_0, a_1), a_2\right) \cdots, a_k \right).
\label{eq:setup}
\end{equation}

Under this transformation, only the initial state changes. A \texttt{Stage} customizes the agent's starting point, either introducing obstacles that require specific skills to tackle or completing early subgoals in advance to shorten the task horizon. For example, in our running task, one \texttt{Stage} executes the following sequence of actions: \texttt{take mug 1}, \texttt{open drawer 1}, \texttt{put mug 1 in drawer 1}, and \texttt{close drawer 1} on the state returned by \texttt{reset()}. This Stage intentionally hides the mug from the agent, forcing the agent to face a more challenging scenario: search for the object first instead of directly reaching for it in plain sight. Conversely, another \texttt{Stage} can simplify the setup by executing \texttt{cleaning the mug} in advance, leaving only the final placement steps.

\paragraph{\texttt{Contract}: rewriting the interaction.}
A \texttt{Contract}, $w_{\mathrm{contract},r}$, is specified by a triplet of transformation maps $r = (f_A, f_T, f_O)$, each defaulting to the identity.
These maps rewrite the action space, transition dynamics, and observation space, respectively:
\begin{equation}
E' = w_{\mathrm{contract}, r}(E) = (\mathcal{S},\; \evocolor{\mathcal{A}'},\; \evocolor{\mathcal{O}'},\; \evocolor{T'},\; R,\; s_0), \quad \text{where } (\evocolor{\mathcal{A}'}, \evocolor{\mathcal{O}'}, \evocolor{T'}) = \big(f_A(\mathcal{A}), f_O(\mathcal{O}), f_T(T)\big).
\label{eq:rule}
\end{equation}

In practice, these transformations enforce action preconditions, augment or mask observations, and attach structured feedback to specific outcomes to steer agent learning. For example, on our running task, one \texttt{Contract} configures $f_O$ to truncate the room description after the first two sentences, requiring the agent to build its spatial representation across several steps; another \texttt{Contract} uses $f_T$ to block the \texttt{clean mug} action if the agent is not holding the mug, forcing the agent to pick up the object first; and a third \texttt{Contract} configures $f_A$ to remove high-level teleport navigation commands, forcing the agent to move and search step by step.

\paragraph{\texttt{Chain}: extending the environment.}
A \texttt{Chain}, $w_{\mathrm{chain},\ell}$, is specified by a pair $\ell = (E_{\mathrm{ext}}, g)$, where $E_{\mathrm{ext}}$ is an additional environment and $g$ is a composition logic.
The composition logic combines the original environment $E$ and $E_{\mathrm{ext}}$ into a composite environment $E'$ exposed through the same interface:
\begin{equation}
E' = w_{\mathrm{chain}, \ell}(E) = \evocolor{(\mathcal{S}', \mathcal{A}', \mathcal{O}', T', R', s_0')}, \quad \text{where } \evocolor{E'} = g(E, E_{\mathrm{ext}}).
\label{eq:link}
\end{equation}

To allow cross-environment combinations, the new spaces are simply the union of the base environments (e.g., $\mathcal{A}' = \mathcal{A} \cup \mathcal{A}_{\mathrm{ext}}$), and $R'$ acts as the new composite reward. The composition logic $g$ is unrestricted, allowing environments to be concatenated, interleaved, or branched dynamically based on intermediate outcomes (see Appendix~\ref{app:link_examples} for examples). 
The composition logic $g$ can either combine both environments from the start, or use the transition function $T'$ to dynamically transition from one environment to the next once a specific condition is satisfied.
For our running environment, one \texttt{Chain} appends \emph{``heat a potato and put it on the countertop''} in the same house, returning success under $R'$ only when both environments are verified. This requires the agent to learn to carry its goal past the point where it would otherwise have stopped.

\paragraph{Composition.} Because all \method components share a standard interface, they compose freely. For example, stacking a \texttt{Stage}, a \texttt{Contract}, and a \texttt{Chain} on our base mug task yields a single composite environment:
\begin{equation}
    E' = w_{\mathrm{chain}, \ell}\left( w_{\mathrm{contract}, r}\left( w_{\mathrm{stage}, \delta}(E) \right) \right).
    \label{eq:composition}
\end{equation}

In this setup, $w_{\mathrm{stage}, \delta}$ initializes the episode with the mug hidden in a drawer to enforce spatial search; $w_{\mathrm{contract}, r}$ truncates the observation to two sentences to evaluate partial observability; and $w_{\mathrm{chain}, \ell}$ appends a follow-up task to test goal persistence. Note that these transformations are non-commutative ($w_1 \circ w_2 \neq w_2 \circ w_1$); the nesting order determines how the resulting environment is constructed and which constraints apply during initialization versus active interaction.

\section{\method for Agent Learning}
\label{sec:methodology}

\subsection{Problem Setup}
\label{subsec:objective}
Given a base environment $E$ supporting a set of base tasks, and a target policy agent $\pi$, our objective is to automatically generate a modified environment $E'$ tailored to a specific task $t$, exposing the unique flaws of $\pi$ to facilitate targeted policy improvement. A single \method{} component is itself policy-agnostic, as Eq.~\eqref{eq:wrapper} defines $w$ as a transformation of the environment alone, allowing the same component to be applied without modification to any policy. Conversely, the selection and parameterization of these components must be conditioned on both the base task $t$ and the observed behavior of the policy $\pi$. We therefore introduce the task-policy-conditioned map $\mathcal{H}$:

\begin{equation}
E' = \mathcal{H}(E, t; \pi) = (w_k \circ w_{k-1} \circ \cdots \circ w_1)(E),
\label{eq:policy_map}
\end{equation}

where each $w_i$ is a customized \method{} component designed to wrap the base environment $E$ and expose the critical weaknesses of $\pi$ on task $t$. Rather than inspecting internal model weights, these components treat the agent as a black box and operate solely on its outputs to generate a steady, corrective training signal.

\subsection{\designer{}}
\label{subsec:generation_loop}

\begin{figure}[t]
    \centering
    \includegraphics[width=0.9\linewidth]{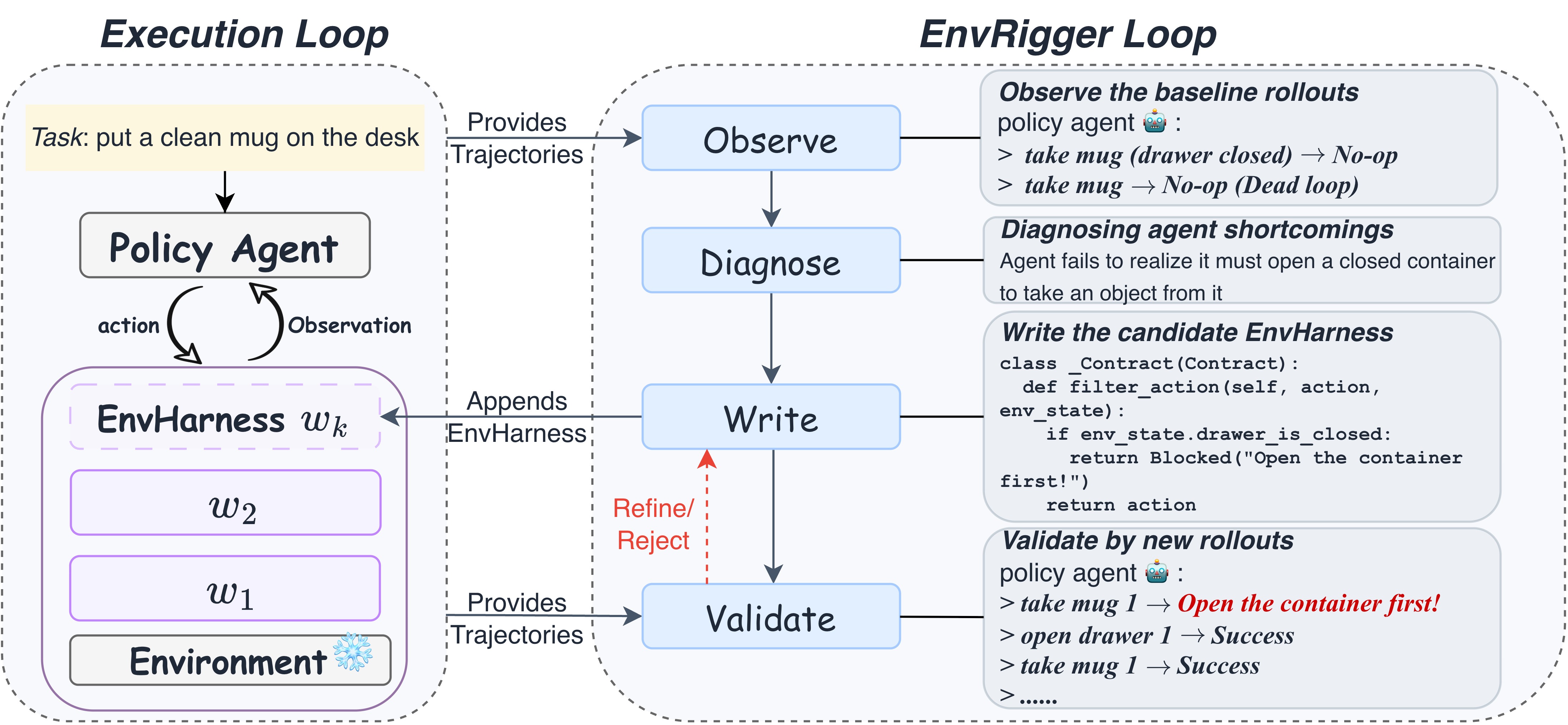}
    \caption{\designer{} generating \method{} components for a target policy based on given task. The \emph{execution loop} on the left runs the policy against the current environment, which is a frozen base environment wrapped by the active \method{} containing accepted components $w_1, \ldots, w_k$, while the resulting rollout trajectories feed the \emph{\designer{} loop} on the right. The \designer{} operates systematically through four distinct stages: Observe, Diagnose, Write, and Validate, where the last two steps form a write-and-validate loop that generates a candidate component, evaluates it on fresh rollouts, and revises it upon failure.}
    \label{fig:generation_loop}
\end{figure}

We introduce \designer{} to realize the task-policy-conditioned map $\mathcal{H}$ of Eq.~\eqref{eq:policy_map}. \designer{} runs $\pi$ in the environment on task $t$, analyzes the resulting trajectories, writes specific \method{} components to customize the environment for this task, and validates the customized environment using fresh policy rollouts.
Candidates from customized environments that provide an appropriate learning signal are accepted, while the \method components of unsuccessful candidates are rejected or revised. Figure~\ref{fig:generation_loop} illustrates this complete workflow, in which \designer{} operates systematically through four distinct stages: Observe, Diagnose, Write, and Validate. To ensure that the initial state mutations introduced by \texttt{Stage} can be reliably reproduced, we assume the base environment supports deterministic resets during the validation phase.

\paragraph{Observe.}
The \designer{} begins by running the policy $\pi$ on the base task $t$ in the current environment to collect and analyze a batch of rollout trajectories. While failures expose the specific weaknesses to be addressed within this task, successes help define the boundaries of these flaws, showing which capabilities are already intact and where they begin to fail.

\paragraph{Diagnose.}
\designer{} analyzes the collected trajectories to identify the root causes of the observed behaviors, focusing on systemic issues such as repetitive action loops, failures in parsing long observations, or misread tool constraints. This diagnosis also determines the customization direction. For a struggling policy, the goal is to scaffold missing steps and simplify the task. Conversely, if the policy achieves a perfect success rate, it indicates the current environment is too forgiving to expose any remaining weaknesses. Under this scenario, the \designer{} diagnoses that the environment must be made harder, shifting the customization to inject more challenging scenarios that force the policy's potential flaws into the open. \designer{} then outputs these findings as a textual diagnosis.

\paragraph{Write.}
Based on the diagnosis, \designer{} synthesizes one or more \method{} components to target the identified flaws. A single flaw may require combining multiple components, such as a \texttt{Stage} that customizes the initial state and a \texttt{Contract} that filters subsequent interactions, emitted together as a candidate set. For example, if the diagnosis reveals that the policy relies on a fragile shortcut that bypasses learning, \designer{} can write a \texttt{Contract} that blocks this action under specific conditions, forcing the policy to explore and master the intended skills.

\paragraph{Validate.}
To evaluate the candidate components from the Write stage, \designer{} wraps the current environment with them to instantiate $E'$ and runs fresh rollouts of $\pi$ on the base task $t$. 
Based on some trajectory metrics, like the success rate and failure distribution, \designer{} analyzes these fresh rollouts to decide between three validation behaviors: accepting the candidate, rejecting those that are unsolvable or non-challenging, or refining candidates with poorly scaled signals.
If the candidate requires refinement, the trajectories and scaling feedback flow back into the Write stage, repeating this Write-and-Validate loop until a candidate is accepted or the revision budget is exhausted. 
All accepted components are ultimately added to the \method{}. 
The exact system prompt and decision-making criteria guiding these validation actions are detailed in Appendix~\ref{app:designer_prompt}.

\section{Experiments}
\label{sec:experiments}

\subsection{Experimental Setup}
\label{subsec:setup}
In this section, we mainly focus on the skill-based learning paradigm, where skills are extracted from environments to improve agent capabilities. We also evaluate the compatibility and performance of our framework under the online reinforcement learning paradigm, which is presented as part of our broader analysis in Section~\ref{sec:analysis}.

\paragraph{Benchmarks and Evaluation.}
We evaluate our framework on five benchmarks spanning four distinct domains: ALFWorld \citep{shridhar2020alfworld} for text-based embodied environments, WebArena \citep{zhou2024webarena} for web interaction, SWE-bench Verified \citep{jimenez2024swebench} for software engineering, and OfficeQA \citep{officeqa} with SpreadsheetBench \citep{ma2024spreadsheetbench} for office automation. We report each benchmark's native metrics, additionally tracking the average steps on SWE-bench Verified as a measure of execution efficiency. Training and evaluation episodes are strictly disjoint on every benchmark, with detailed split configurations provided in Appendix~\ref{app:splits}. 

Importantly, \designer and the policy agent utilize the same model backbone on each benchmark: Gemini-3.1-Flash-Lite for ALFWorld and WebArena, and Gemini-3.5-Flash elsewhere, ensuring that performance gains do not stem from distilling a stronger external model. On each training set, \designer executes the optimization loop of Section~\ref{sec:methodology} to generate \method-customized environments. We then extract skills from trajectories collected in these environments following ReasoningBank \citep{ouyang2025reasoningbank}, and evaluate the skill-equipped policy agent on the held-out instances.
We exclude the \textit{Chain} component from this automated pipeline because it is difficult for \designer{} to observe the internal states of joined environments. Instead, we analyze the effect of chaining separately in Section~\ref{sec:analysis}.

\paragraph{Baseline Methods.}
\label{subsec:baselines}
We compare \method against four skill sources: \textbf{No Skills} (the frozen policy agent alone), \textbf{Original Envs} (skills from original environments to isolate reshaping effects), and \textbf{GenEnv} \citep{guo2025genenv}, \textbf{VeriEnv} \citep{chae2026safe}, or \textbf{SWE-smith} \citep{yang2026swe} for their respective benchmarks. While these baseline generators are domain-specific, \method applies generically across all domains via a unified interface, including the office automation environments where no generation baseline exists. To ensure a fair comparison, all baselines share the same seed instances, environment count, extraction pipeline, and policy model. \designer operates strictly on training episodes under the same oracle verification access, and each evaluation instance is attempted once. Baseline details are in Appendix~\ref{app:baselines}.

\subsection{Main Results}
\label{subsec:main_results}

Table~\ref{tab:main_results} and Table~\ref{tab:main_results_swe} summarize the primary results across five benchmarks. Based on these evaluations, we present the following key observations.

\begin{table}[ht]
\centering
\caption{Performance of agents equipped with skills extracted from different environment sources on ALFWorld and WebArena. All numbers are the mean over three independent runs, with
standard deviations as \textcolor{gray}{gray subscripts}. Higher is better for every metric, and the last row reports the improvement of \method{} Envs over Original Envs. ``--'' denotes that the method is benchmark-specific and cannot be applied to the other domain.}
\label{tab:main_results}
\resizebox{\textwidth}{!}{
\begin{tabular}{l ccc c ccccc}
\toprule
\multirow{2}{*}{\textbf{Skill Source}} & \multicolumn{3}{c}{\textbf{ALFWorld}} & & \multicolumn{5}{c}{\textbf{WebArena}} \\
\cmidrule{2-4} \cmidrule{6-10}
& \textbf{In-Dist} & \textbf{OOD} & \textbf{Avg.} & & \textbf{Reddit} & \textbf{Shopping} & \textbf{Shop Admin} & \textbf{GitLab} & \textbf{Avg.} \\
\midrule
No Skills              & 62.6\sd{1.7} & 60.7\sd{5.2} & 61.7\sd{3.4} & & 39.6\sd{2.3} & 35.2\sd{3.3} & 44.1\sd{2.3} & 35.8\sd{8.4} & 38.7\sd{2.3} \\
Original Envs          & 63.3\sd{2.8} & 61.4\sd{4.3} & 62.4\sd{3.4} & & 38.7\sd{9.7} & 35.2\sd{1.3} & 44.6\sd{3.0} & 35.4\sd{4.0} & 38.5\sd{3.1} \\
GenEnv                 & 63.3\sd{1.2} & 61.9\sd{2.7} & 62.6\sd{1.9} & & --           & --           & --           & --           & --           \\
VeriEnv                & --           & --           & --           & & 39.6\sd{4.2} & 30.2\sd{0.0} & 49.7\sd{2.4} & \textbf{38.9}\sd{5.6} & 39.6\sd{1.4} \\
\method{} Envs         & \textbf{66.2}\sd{0.3} & \textbf{70.4}\sd{2.3} & \textbf{68.3}\sd{1.3} & & \textbf{40.6}\sd{4.7} & \textbf{37.4}\sd{0.3} & \textbf{50.8}\sd{1.5} & 37.7\sd{3.1} & \textbf{41.6}\sd{1.8} \\
\midrule
\rowcolor{black!5}
Improvement & +2.9 & +9.0 & +5.9 & & +1.9 & +2.2 & +6.2 & +2.3 & +3.1 \\
\bottomrule
\end{tabular}
}
\end{table}
\begin{table}[h]
\centering
\caption{Performance of agents equipped with skills extracted from different environment sources on  SWE-bench Verified, OfficeQA, and
SpreadsheetBench. Standard deviations are
\textcolor{gray}{gray subscripts}.  ``--'' denotes that the method is benchmark-specific and cannot be applied to the other domain.}
\label{tab:main_results_swe}
\resizebox{\textwidth}{!}{
\begin{tabular}{l cc c cc c cc}
\toprule
\multirow{2}{*}{\textbf{Skill Source}} & \multicolumn{2}{c}{\textbf{SWE-verified}} & & \multicolumn{2}{c}{\textbf{OfficeQA}} & & \multicolumn{2}{c}{\textbf{SpreadsheetBench}} \\
\cmidrule{2-3} \cmidrule{5-6} \cmidrule{8-9}
& \textbf{SR ($\uparrow$)} & \textbf{Average Step ($\downarrow$)} & & \textbf{EM ($\uparrow$)} & \textbf{F1 ($\uparrow$)} & & \textbf{Pass@1 ($\uparrow$)} & \textbf{Mean Score ($\uparrow$)} \\
\midrule
No Skills            & 47.67\sd{0.93}          & 53.58\sd{2.93}          & & 54.23\sd{2.84}          & 55.77\sd{2.98}          & & 46.44\sd{0.15}          & 61.32\sd{0.37}          \\
Original Envs        & 49.88\sd{2.59}          & 55.01\sd{1.69}          & & 54.40\sd{1.84}          & 55.77\sd{1.59}          & & 45.88\sd{1.19}          & 61.47\sd{0.59}          \\
SWE-smith            & 50.12\sd{1.74}          & 54.72\sd{2.03}          & & --             & --             & & --                      & --                      \\
\method{} Envs       & \textbf{52.58}\sd{2.72} & \textbf{49.61}\sd{2.49} & & \textbf{56.20}\sd{2.34} & \textbf{57.73}\sd{2.29} & & \textbf{49.15}\sd{0.36} & \textbf{62.48}\sd{0.27} \\
\midrule
\rowcolor{black!5}
Improvement& +2.70          & +5.40          & & +1.80          & +1.96          & & +3.27          & +1.01          \\
\bottomrule
\end{tabular}
}
\end{table}

\paragraph{\method delivers consistent gains where static environments cannot.}
Skills acquired in environments customized by \method consistently outperform those extracted from original environments on every benchmark, yielding up to a 9.0-point improvement on ALFWorld. Conversely, extracting skills from static base environments can actually degrade performance; for instance, on SpreadsheetBench, skills from unmodified environments fall below the no-skill baseline, while on SWE-bench Verified, they lengthen execution trajectories. Because static environments only allow the agent to practice behaviors it already executes, they fail to address its specific limitations, often retrieving redundant or suboptimal skills. In contrast, the write-and-validate loop of \designer only commits environment components verified by fresh policy trajectories, ensuring that \method consistently improves upon the no-skill baseline across all benchmarks.

\paragraph{\method generalizes across domains through a domain-agnostic interface.}
The underlying interface protocol, \designer loop, and the skill extraction pipeline apply consistently across all five benchmarks, requiring only domain-specific prompt templates to adapt to different environments. By contrast, specialized generation baselines are constrained to their specific target benchmarks (indicated by dashes for inapplicable domains in Tables~\ref{tab:main_results} and \ref{tab:main_results_swe}). Nevertheless, \method{} consistently outperforms these specialized baselines wherever they can be applied. On ALFWorld, \method skills surpass GenEnv by 5.7 points on average and by 8.5 points in out-of-distribution settings, where generic instance generation merely increases repetitive practice without addressing policy weaknesses. On SWE-bench Verified, \method outperforms the purpose-built SWE-smith by 2.46 points in success rate while requiring 5.11 fewer execution steps per episode. Targeting diagnosed vulnerabilities through a unified interface thus proves superior to merely scaling up the quantity of training episodes through domain-specific generation.

\paragraph{\method improves efficiency by repairing wasteful behaviors.}
On SWE-bench Verified, skills extracted from \method-customized environments reduce the average steps per episode from 53.6 to 49.6, whereas skills from unmodified environments actually increase it to 55.0. This efficiency gain directly correlates with the specific diagnostics from \designer: targeted \texttt{Contracts} and \texttt{Stages} designed to disrupt repetitive action loops and filter verbose observations successfully shorten execution trajectories.

\section{Analysis}
\label{sec:analysis}

We analyze \method along five dimensions: its compatibility as a training signal for reinforcement learning, its scaling efficiency compared to standard dataset expansion, its transferability across different policy model families and strengths, the unique value of the \texttt{Chain} component on long-horizon environments, and the capability of the \designer to accept explicit, user-defined target constraints. Additional analyses are reported in Appendix~\ref{app:add_analysis}.

\paragraph{\method enables better RL.}
\begin{wraptable}{r}{0.55\textwidth}
\vspace{-4mm}
\centering
\small
\setlength{\tabcolsep}{3pt}
\renewcommand{\arraystretch}{1.1}
\begin{tabular}{l ccc cc}
\toprule
& \multicolumn{3}{c}{\textbf{ALFWorld}} & \multicolumn{2}{c}{\textbf{WebShop}} \\
\cmidrule(lr){2-4}\cmidrule(lr){5-6}
\textbf{Training Set} & \textbf{In-Dist} & \textbf{OOD} & \textbf{Avg.} & \textbf{Score} & \textbf{SR} \\
\midrule
Original Envs      & 81.4 & \textbf{89.6} & 85.5 & 75.6 & 66.0 \\
\method Envs       & \textbf{87.9} & 88.8 & \textbf{88.4} & \textbf{79.2} & \textbf{67.4} \\
\bottomrule
\end{tabular}
\vspace{1mm}
\caption{Reinforcement learning on ALFWorld and WebShop, comparing policies trained on the original environments and on \method environments. ALFWorld is scored by success rate on in-distribution and held-out instance types; WebShop reports environment score and success rate.}
\label{tab:rl_results}
\vspace{-2mm}
\end{wraptable}

Beyond skill-based learning, we investigate whether \method-customized environments can provide active training signals in online reinforcement learning.
We perform this analysis on ALFWorld and WebShop \citep{yao2022webshop} using Qwen3-8B-base \citep{yang2025qwen3} as the policy, optimized via Group Relative Policy Optimization (GRPO) \citep{shao2024deepseekmath}; training details are provided in Appendix~\ref{sec:appendix_rl_details}. 
We train two distinct policies: one trained solely on the original static environments, and one trained entirely on environments reshaped by \method, evaluating both on the same held-out instances. Table~\ref{tab:rl_results} reports the results. Training on \method environments consistently improves policy performance, outperforming the baseline trained solely on original environments on three out of four metrics. Specifically, on ALFWorld, \method Envs achieves an in-distribution success rate of 87.9 compared to 81.4 for original environments. On WebShop, it achieves a higher score of 79.2 (versus 75.6) and success rate of 67.4 (versus 66.0). Although there is a slight, negligible trade-off in the ALFWorld OOD success rate (88.8 versus 89.6), the overall results underscore a fundamental advantage: the reshaped environments are not merely auxiliary data but provide a highly effective, independent optimization signal for online policy learning.


\paragraph{\texttt{Chain} enables efficient long-horizon task solving.}

\begin{wraptable}{r}{0.6\textwidth}
\centering
\small
\setlength{\tabcolsep}{3.5pt}
\vspace{-6mm}
\begin{tabular}{lcc}
\toprule
\textbf{} & \textbf{SR (\%)} $\uparrow$ & \textbf{AS} $\downarrow$ \\
\midrule
No Skills & 47.67 & 53.58 \\
Original Envs & 49.88 & 55.01 \\
\method{} (\texttt{Stage}/\texttt{Contract} Only) & 52.58 & 49.61 \\
\method{} (\texttt{Chain} Only) & 49.63 & \textbf{41.96} \\
\midrule
Combined Skills (\texttt{Stage}/\texttt{Contract} + \texttt{Chain}) & \textbf{54.30} & 43.12 \\
\bottomrule
\end{tabular}
\vspace{1mm}
\caption{Performance on long-horizon environments. SR stands for Success Rate, and AS represents Average Steps.}
\label{tab:link_results}
\vspace{-2mm}
\end{wraptable}

Real-world applications often require agents to operate over extended horizons. The \texttt{Chain} component addresses this by joining two randomly paired base environments into a single, extended episode. To isolate its effect, this pairing operates independently of the autonomous \designer{} loop. We evaluate the extracted skills on standard single-environment test instances, reporting Success Rate (SR) and Average Steps (AS) in Table~\ref{tab:link_results}.

Skills from \texttt{Chain} environments yield substantial efficiency gains, reducing AS from 53.58 to 41.96. Their standalone SR (49.63) is marginally below the 49.88 baseline. This aligns with their stringent training condition—where success requires solving both halves—which prioritizes long-term goal preservation over short-task maximization. Combining both skill sets (\texttt{Stage}/\texttt{Contract} + \texttt{Chain}) achieves the best of both worlds: the highest SR (54.30) and excellent efficiency (43.12 AS), demonstrating highly complementary behaviors. Representative skills are in Appendix~\ref{app:link_skills}.

\paragraph{\method enables efficient environment scaling.}
\begin{wrapfigure}{r}{0.5\textwidth}
\centering
\vspace{-8mm} 
\includegraphics[width=\linewidth]{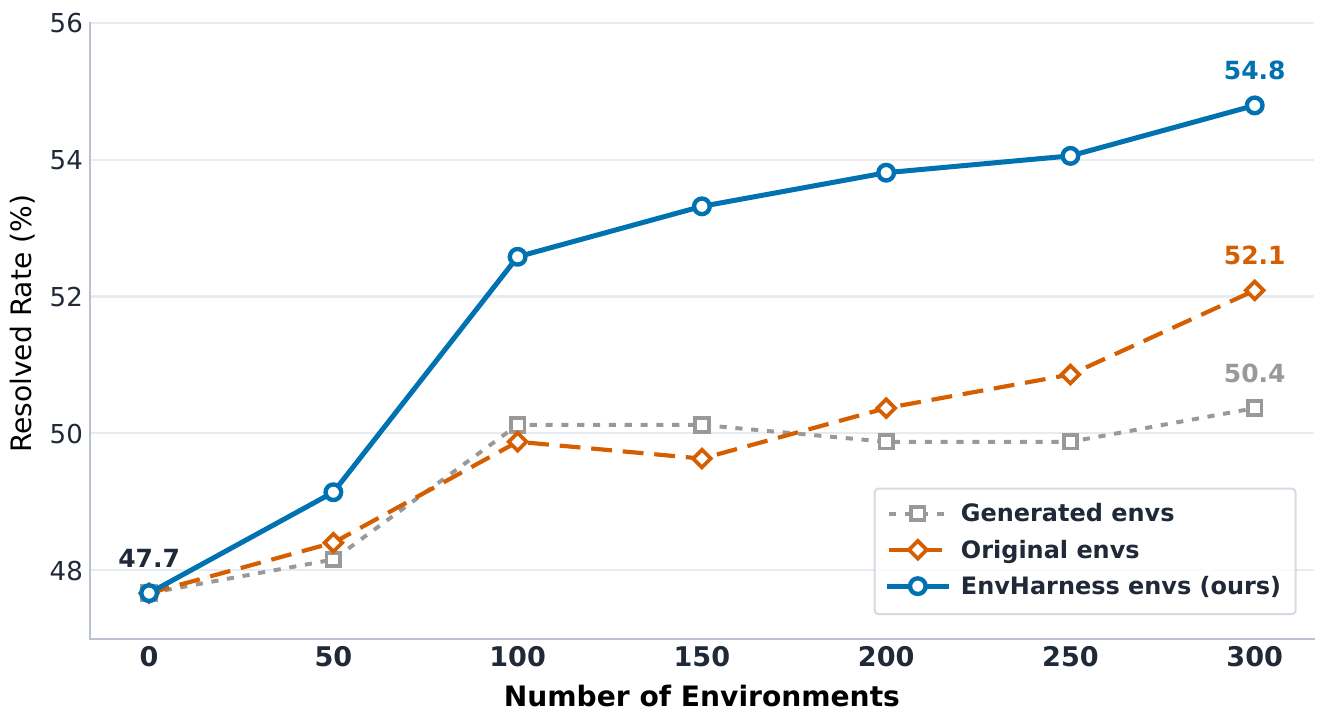}
\caption{Environment scaling on SWE-bench Verified. All three sources supply the same number of environments and feed the same extraction and retrieval protocol.}
\label{fig:coevo}
\vspace{-2mm}
\end{wrapfigure}

Environment scaling evaluates performance as the available training environments expand, comparing three allocation strategies under an identical budget: \method{} environments, unmodified benchmark environments, and SWE-smith generated environments. Holding the policy model, environment budget, and skill retrieval protocol fixed, each batch of 50 environments yields one skill bank (alternating between 2 and 3 skills per bank, totaling 15 skills at 300 environments). Crucially, while both baselines draw environment batches independently of the learner, \method{} synthesizes each batch specifically targeting the policy equipped with previously accumulated skills, enabling the environments and the policy to co-evolve. As shown in Figure~\ref{fig:coevo}, \method{} climbs from 47.67 to 54.79 (a 7.12-point gain) and maintains an upward trajectory at 300 environments. In contrast, the same budget yields only 52.13 on original environments and 50.37 on generated ones. This performance gap confirms that targeting the learner's current capability boundary is fundamentally more effective than unconditioned environment scaling. Representative skills from each round are provided in Appendix~\ref{app:coevo_skills}.

\begin{wrapfigure}{r}{0.55\textwidth}
\centering
\vspace{-12mm}
\includegraphics[width=\linewidth]{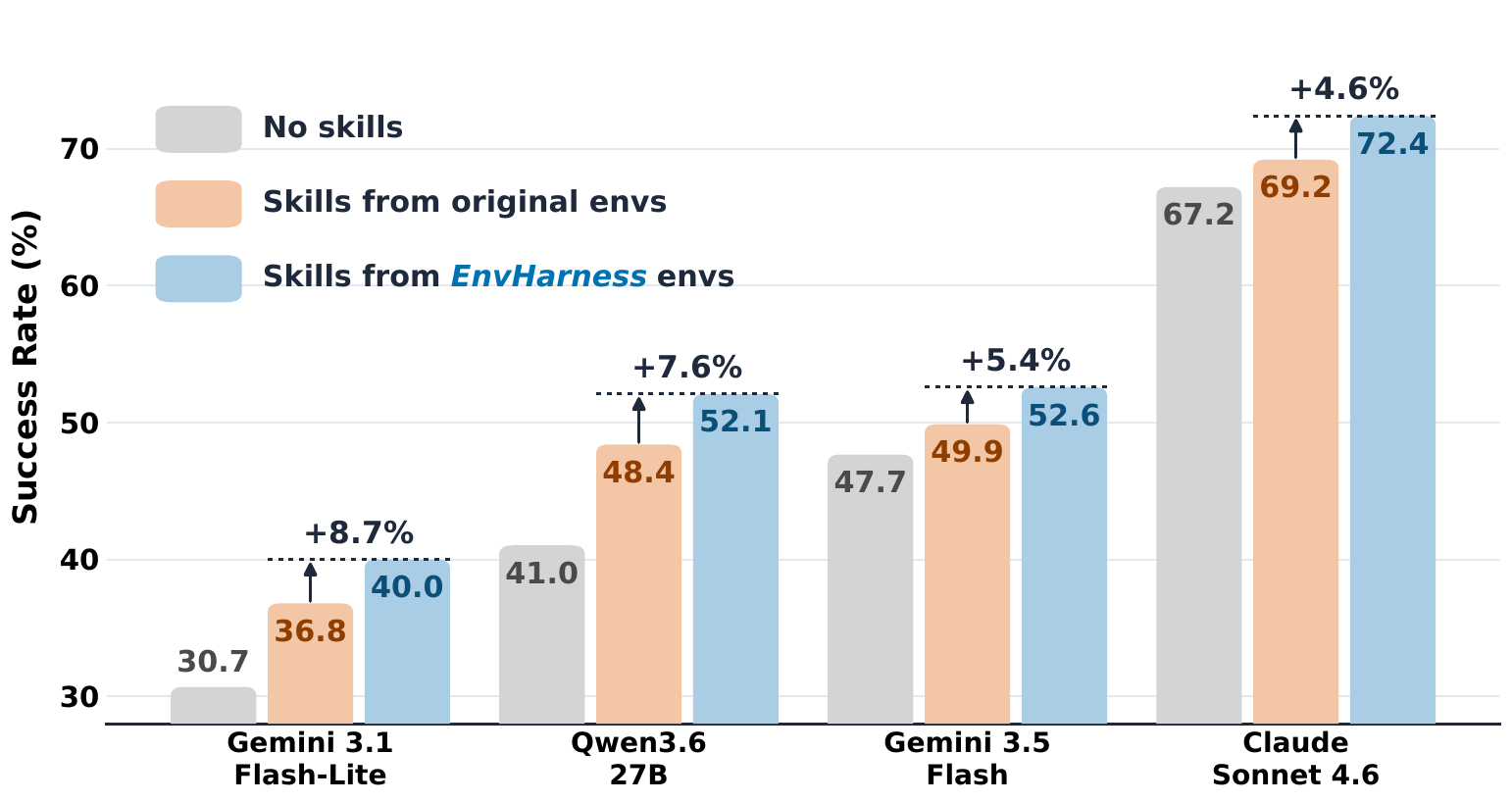}
\caption{Cross-model results on SWE-bench Verified. Each group represents one policy model, ordered from weakest to strongest. Bars show success rates with no skills, with skills from unmodified environments, and with skills from \method environments. Percentages represent relative gains over original environments.}
\label{fig:crossmodel}
\vspace{-2mm}
\end{wrapfigure}

\paragraph{\method generalizes across different LLM backbones.}
To evaluate generalizability, we test four distinct models on SWE-bench Verified: Gemini 3.1 Flash-Lite, Qwen3.6 27B \citep{qwen3.6-27b}, Gemini 3.5 Flash, and Claude Sonnet 4.6 \citep{anthropic2026claude46sonnet}. These span open-weight and proprietary architectures across a wide capability spectrum. In each setting, we use the same model backbone for both the target policy and the \designer{} to keep the setup consistent, while the extraction pipeline and protocol remain unchanged.

Figure~\ref{fig:crossmodel} shows the results. \method skills outperform real environment skills on all four policies, by 2.7 to 3.7 absolute points, even though the skill-free success rates span a broad range from 30.7 to 67.2. The size of this gain is largely independent of how strong the underlying policy is: the customization loop neither breaks down on the weakest model nor saturates on the strongest, and the same pipeline, prompts, and acceptance criteria are used throughout. What the policy's capability level appears to change is the content of the diagnoses rather than the applicability of the loop. We note that skills of either kind help the two weakest policies most relative to no skills at all (+9.3 and +11.1 points for \method, and +6.1 and +7.4 for unmodified environments, versus under 5.5 points for the two strongest).

\paragraph{\method produces environments on demand.}
While the \designer loop autonomously identifies training targets via behavioral diagnosis in standard settings, the same machinery can readily accept explicit, user-defined constraints. 
We evaluate two classes of constraints. The first is a quantitative target (success rate or average steps) for an objective performance metric in Appendix~\ref{app:add_analysis}.
The second constraint targets capability weaknesses described in natural language. For example, given the weakness below, the \designer{} generates a \texttt{Contract} that rejects code submissions unless tests are run. This forces the agent to verify its fixes. From the resulting trajectories, we distill the skill shown below. Rather than overfitting to a single task, the skill combines a general principle with actionable steps, perfectly matching our requirement for a skill~\citep{yang2026skillopt}.

\begin{weaknessbox}
The policy submits a patch without running the failing test, so the fix stays unverified.
\end{weaknessbox}

\begin{componentbox}{Generated component (Contract, $f_T$ axis)}
class _Contract(Contract):
    def modify_transition(self, action, response, env_state):
        cmd = bash_command(action)
        if "pytest" in cmd or "runtests.py" in cmd:
            env_state.extras["ran_tests"] = True
        if is_submission(cmd) \
                and not env_state.extras.get("ran_tests"):
            return failed(response,
                "githook: pre-commit hook 'verify-tests' "
                "failed. Run the test suite before submitting.")
        return response
\end{componentbox}

\begin{skillbox}{Verification-Driven Development Loop}
\skillfield{Description} Whenever a code change is made to fix a bug or implement a feature, especially where the test suite needs setup or configuration. \\
\skillfield{Content} Before finalizing any change, run the relevant test suite to confirm the failure exists, then run it again after the patch to verify the fix, initializing the environment first when needed.
\end{skillbox}

\section{Related Work}
\subsection{Environment Scaling}
Environment scaling has emerged as a research direction that supplies agents with more environments to learn from \citep{huang2025environment, xi2025agentgym}. Existing efforts scale environments in various forms, including simulating environments and their feedback with an LLM \citep{guo2025genenv, zala2024envgen, wang2025llms}, up to world models that simulate or synthesize whole families of agentic environments \citep{zuo2026qwen, wang2026agent}, synthesizing executable environments programmatically \citep{chae2026safe, song2026envscaler, tang2026phoneworld, sun2026swe, dong2026agent}, and synthesizing new task instances inside an existing benchmark \citep{yang2026swe, pan2024training}. Beyond producing more tasks, another line adapts what the environment presents to the learner, from curriculum generation in reinforcement learning \citep{jiang2021prioritized, dennis2020emergent, wang2019paired, liu2026spade} to hand-designed corrective feedback and reward shaping \citep{lu2025don}. Different from previous works that rely on benchmark-specific pipelines or hand-designed curricula, \method reshapes an existing environment through one interface shared across benchmarks, conditions the reshaping on the diagnosed weaknesses of the current policy, and leaves tasks and verifiers untouched.

\subsection{Self-Evolving Agent}
Self-evolving agents improve themselves from their own experience without additional human supervision \citep{fang2025comprehensive}. Existing efforts evolve different parts of the agent, including prompts and reflections \citep{shinn2023reflexion, madaan2023self}, skill and workflow libraries \citep{wang2023voyager, wang2024agent, xia2026skillrl, yang2026skillopt, xia2026metaclaw, huang2026raw}, experience memory distilled from past trajectories \citep{zhao2024expel, ouyang2025reasoningbank}, the model weights through self-generated rewards or self-proposed tasks \citep{yuan2024self, huang2025r, zhao2026absolute, xia2025agent0, he2025visplay, huang2026g}, and recently the agent harness itself, rewritten and tested around a frozen model \citep{lee2026meta}. Different from these methods that evolve the agent while the world it learns from stays fixed, \method reshapes the environment itself against the diagnosed weaknesses of a frozen policy.
\section{Conclusion}
We introduce \method{}, a programmable layer that turns a static, existing environment into a controllable one. \method{} wraps a frozen benchmark with three plug-in components, Stage, Contract, and Chain, and reshapes it entirely through the standard \texttt{reset}/\texttt{step} interface, making it possible to isolate a skill, extend a task's horizon, or calibrate difficulty in environments that were never built for any of these purposes. Since \method{} never touches internal code, a single implementation works seamlessly across different domains. Furthermore, by leaving the original tasks unchanged, every reshaped environment safely retains its trusted, human-built verifier. To fully automate this customization, we introduce \designer{}, an autonomous loop that diagnoses policy weaknesses from execution trajectories and synthesizes targeted \method{} components to provide precise learning signals. This reframes environment construction as a wrapping problem rather than an authoring one, and suggests a practical pathway toward scalable environment supply for agent learning. We present future directions and limitations in Appendix~\ref{app:future} and Appendix~\ref{app:limitation}.

\bibliography{main}


\clearpage
\appendix


\vspace*{1em} 
\noindent{\Large \textbf{Contents of Appendix}} 
\vspace{0.5em} 
\hrule height 0.8pt 
\vspace{1em} 

\startcontents[appendix]
\printcontents[appendix]{l}{1}{\setcounter{tocdepth}{2}}

\clearpage

\section{The \designer{} Prompt}
\label{app:designer_prompt}

\paragraph{Naming.}
The released code predates the terminology of this paper. The three
component types appear there as the classes \texttt{Setups},
\texttt{Rules}, and \texttt{Link}, and a candidate emitted by the designer
carries the fields \texttt{rules\_code} and \texttt{in\_env\_actions}.
Table~\ref{tab:naming} maps the two, and the rest of this appendix uses the
names of the paper.

\begin{table}[h]
\centering
\small
\begin{tabular}{lll}
\toprule
\textbf{Paper} & \textbf{Code} & \textbf{Emitted field} \\
\midrule
Stage    & \texttt{Setups} & \texttt{in\_env\_actions} \\
Contract & \texttt{Rules}  & \texttt{rules\_code} \\
Chain    & \texttt{Link}   & --- \\
\bottomrule
\end{tabular}
\caption{Component names in the paper and in the release.}
\label{tab:naming}
\end{table}

The prompt below is the part shared by every benchmark. Each benchmark
appends a short block of its own detailing the tools its bridge exposes,
the fields of its \texttt{env\_state} view, and its domain-specific
constraints; those blocks are in the code release.

\begin{promptbox}

\begin{lstlisting}[style=prompt]
You are the Environment Designer for an agent benchmark. Your job is to reshape the environment so the Policy agent gets the right training signal. You emit a Candidate with two independent levers:

1. `rules_code` -- a Python class `_Rules(Rules)` overriding up to three per-step hooks: filter_action (A axis: transform or Block an action), modify_transition (T axis: transform the env's response), and filter_observation (O axis: transform what the Policy sees). All hooks default to pass-through; the class is loaded fresh per episode.

2. `in_env_actions` -- a list of tool calls the framework REPLAYS through env.step() before the Policy starts. This is the S0 (initial-state) mechanism: instead of writing code, you write a trajectory the environment walks for you.

The two levers compose freely: S0-only, hooks-only, or both (e.g. seed a state via in_env_actions, then block the easy escape via filter_action). The R axis is not exposed: success is the benchmark's own verdict, so reshaping reward cannot move the eval metric. Hooks may read the env_state schema provided each turn; import only the standard library.

PITFALL -- do not make the task unsolvable. A mutation that makes success impossible is not a difficulty increase; SR=0 from impossibility is exactly as useless as SR=1 from triviality. Signals that a prior mutation was unsolvable: most rollouts end in timeout, or SR=0 with failures pointing at the action axis. On these signals your next proposal must REVERSE or loosen the offending restriction -- stacking more bans cannot climb into the band. Prefer subtle, narrow perturbations (one op, one obs key) over sweeping bans.

BASELINE -- before your first proposal you see K unmutated rollouts on this task: success rate, per-rollout outcomes, and sample trajectories. Read it for three things: WHETHER the Policy can solve the task at all (if baseline SR is ~0, "make it harder" is nonsensical -- scaffold it easier or skip); HOW it solves it (a 4-step solution leaves less headroom than a 30-step one -- match perturbation magnitude to that headroom); and WHICH parts of the env it actually relies on (perturbing commands it never uses is irrelevant). Treat the baseline as raw data, not as a hint: decide direction and magnitude yourself.

REFINE -- after K rollouts of your candidate you decide ACCEPT / REFINE / REJECT, referencing rollout statistics (SR over K runs, failure distribution, timeout count), never a single trace. When refining, ask: did this mutation move SR toward the target band? If yes, the perturbation TYPE is right -- keep the working hooks verbatim and adjust only the magnitude (loosen if overshot, tighten if undershot); do not discard code that paid K rollouts of signal to establish. If no, start over with a different perturbation type.

Your operating mechanism is fixed. The OBJECTIVE that tells you what to optimize is provided each turn; per-benchmark constraints are appended as experiment-specific instructions.
\end{lstlisting}
\end{promptbox}

\section{Distinct Differences from Related Co-Evolution and Synthesis Frameworks}
\label{app:detailed_differences}

To clearly position \method{} within the literature on adaptive environment design and co-evolution, we highlight the core differences between our framework and three closely related paradigms: generative co-evolution, adaptive configuration engines, and programmatic environment synthesis.

\paragraph{1. Comparison against Generative Co-Evolution (e.g., GenEnv).}
\begin{itemize}[leftmargin=1.5em, itemsep=0.2em]
    \item \textbf{Their Approach:} \texttt{GenEnv}~\citep{guo2025genenv} uses an LLM as a generative simulator to dynamically generate transitions, observations, and success signals on the fly.
    \item \textbf{The Limitation:} Relying on LLMs to simulate physical transitions and verify tasks inherently introduces hallucinations and evaluation drift, which compromises the mathematical validity of the benchmark.
    \item \textbf{Our Solution:} \method{} leaves the base environment, its native transition function $T$, and trusted, human-built verifiers completely \textit{frozen}. All customizations are applied non-invasively at the standard interface boundary via Stage, Contract, and Chain wrappers, ensuring 100\% deterministic transition logic and high-trust evaluation integrity.
\end{itemize}

\paragraph{2. Comparison against Adaptive Configuration Engines (e.g., EnvGen).}
\begin{itemize}[leftmargin=1.5em, itemsep=0.2em]
    \item \textbf{Their Approach:} \texttt{EnvGen}~\citep{zala2024envgen} generates and adapts environment configurations (e.g., changing maps or terrain files) inside the simulator core.
    \item \textbf{The Limitation:} Modifying a simulator's internal code or asset configurations is highly benchmark-specific, requiring deep, manual engineering for every new domain and risking corrupted state logic.
    \item \textbf{Our Solution:} \method{} is entirely benchmark-agnostic. While adding a new environment requires implementing a one-time lightweight wrapper bridge (the standard \texttt{ActionableEnv} interface), the core co-evolution loop and the environment designer require no further modification. Once the interface is established, \method{} automatically generates and applies customized components to any benchmark (e.g., ALFWorld, WebArena, or SWE-bench) without any manual, task-specific configurations.
\end{itemize}

\paragraph{3. Comparison against Programmatic Environment Synthesis (e.g., Agent-World).}
\begin{itemize}[leftmargin=1.5em, itemsep=0.2em]
    \item \textbf{Their Approach:} \texttt{Agent-World}~\citep{dong2026agent} programmatically synthesizes executable toolsets, databases, and tasks from scratch to scale up training instances.
    \item \textbf{The Limitation:} Synthesizing fully executable environments from scratch incurs massive engineering overhead, and the generated tools may contain logic errors that stall agent training.
    \item \textbf{Our Solution:} Rather than building new environments from scratch, \method{} non-invasively \textit{repurposes} existing, highly trusted, and established benchmarks. By diagnosing policy weaknesses under a base task $t$, our designer automatically constructs target-oriented challenge wrappers. This significantly reduces computational and engineering overhead while preserving the rigorous grading criteria of established research baselines.
\end{itemize}

\section{Interface Protocol and Design Patterns}
\label{sec:interface}

\begin{figure}[t]
\centering
\includegraphics[width=\linewidth]{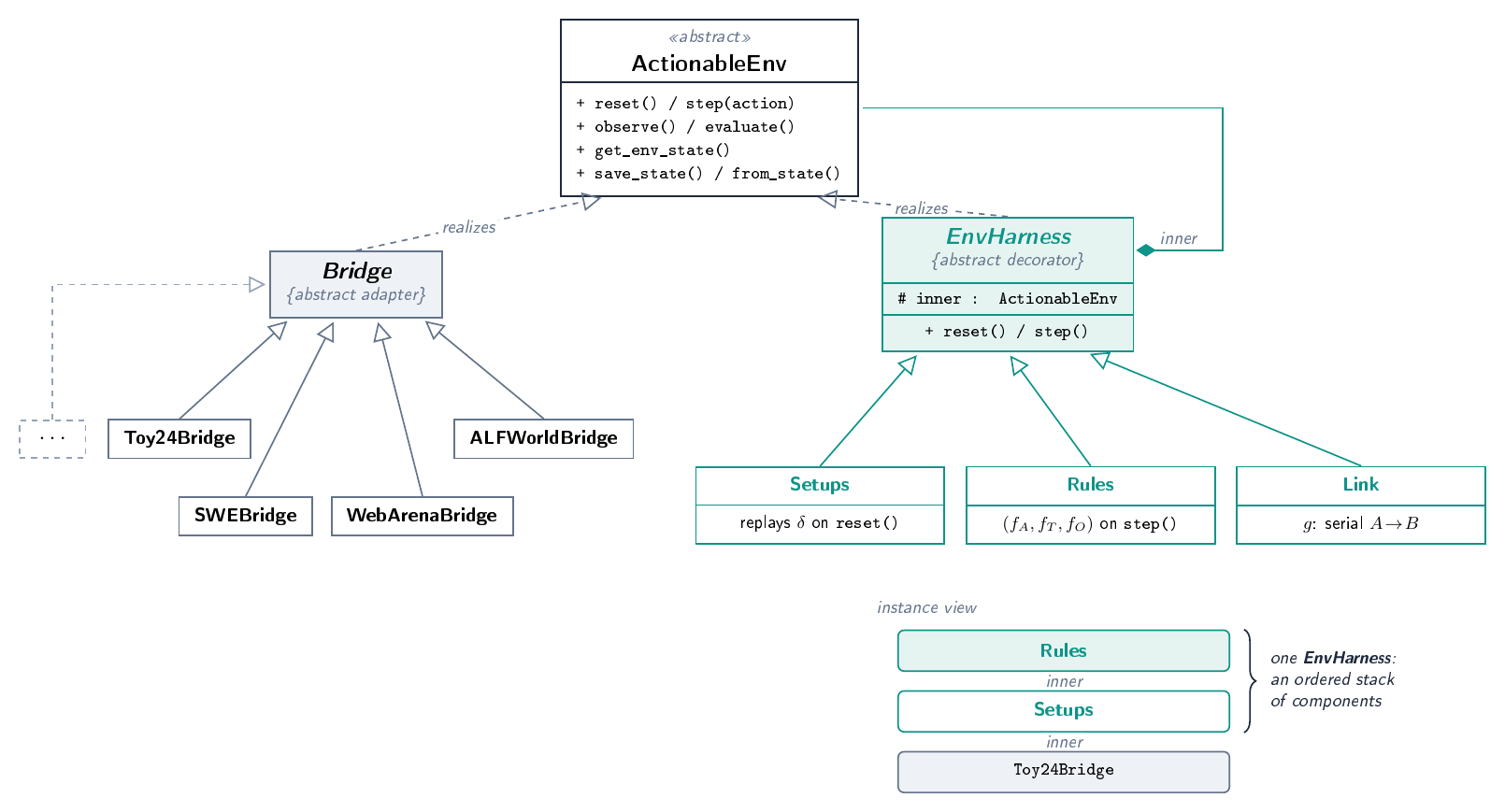}
\caption{Class structure and instance structure of the framework.
Top left, seven Bridges adapt heterogeneous native runtimes to the abstract \texttt{ActionableEnv} contract, and the dashed box marks that further Bridges plug in the same way. Top right, \texttt{EnvHarness} is the abstract decorator over the same contract, and the three shipped components derive from it. Bottom, the instance view shows one \method assembled as an ordered stack of components over a Bridge. Each layer holds the next as \texttt{inner} and, by construction, never accesses the runtime beneath the interface. Any class honoring the contract is a valid component, and the family is open to extension.}
\label{fig:uml}
\end{figure}

\method is built around a single design commitment: \emph{every benchmark, and every transformation of a benchmark, presents the same interface}. A policy, an orchestrator, or a component layer programs against one abstract type and cannot distinguish a raw benchmark from a benchmark wrapped in an arbitrary stack of \method components. Figure~\ref{fig:uml} summarizes the resulting class structure. This section describes the three levels of the protocol: the universal environment interface (\S\ref{sec:actionable-env}), the per-benchmark \emph{Bridges} that adapt heterogeneous runtimes to it (\S\ref{sec:bridges}), and the component layer that assembles one \method per environment (\S\ref{sec:envharness}).

\subsection{\texttt{ActionableEnv}: The Interactable-Environment Interface}
\label{sec:actionable-env}

At the base of the framework is \texttt{ActionableEnv}, an abstract class that fixes the contract every environment must satisfy. It consists of two groups of methods. The first group is a Gymnasium-style interaction loop with typed, validated data contracts. \texttt{reset(seed, options)} initializes an episode. \texttt{step(action)} consumes an \texttt{Action} (a tool name plus JSON-serializable keyword arguments) and returns an \texttt{EnvResponse}, a Pydantic wrapper around the Gymnasium 5-tuple (observation, reward, terminated, truncated, info). \texttt{evaluate()} returns a terminal \texttt{EvaluationResult}, and \texttt{observe()} returns a fresh \texttt{Observation} of the current state. \texttt{observe()} is deliberately separated from \texttt{reset()}: a component may mutate the environment \emph{after} reset returns but \emph{before} the policy acts, and \texttt{observe()} lets the outer layer re-read the world without paying for another reset. Finally, \texttt{get\_env\_state()} exposes a \emph{runtime-safe view} of the internal state. This view consists of plain data with no Docker handles, browser pages, or sockets, and it is the only state that component hooks are permitted to read. This restriction is what makes component code portable: the same hook that runs against an in-memory puzzle also runs against a containerized repository, because neither ever touches the runtime beneath the state view.

The second group makes persistence environment-owned. \texttt{save\_state()} returns a JSON-serializable dictionary, and the classmethod \texttt{from\_state(dict)} reconstructs an instance from it. The contract intentionally does not prescribe \emph{what} to save. Pure in-memory environments serialize their full live state, while environments whose runtime cannot be cheaply cloned (containers, browsers, game engines) save only their reset arguments and accept that restoration is valid at episode boundaries. Each concrete class registers a stable string tag through a registry decorator, so saved stacks can be reconstructed without embedding import paths in checkpoint files.

The interface also declares optional capabilities with safe defaults: a per-step dense reward hook \texttt{step\_reward(step\_info)} (non-fatal by contract, where exceptions are recorded but never episode-terminating), a \texttt{notify\_replay\_complete()} callback that lets an environment rewind per-episode bookkeeping after a state-preparation replay, task enumeration via \texttt{list\_tasks()}, and \texttt{close()} for releasing external runtime resources.

\subsection{Bridges: Adapting Heterogeneous Benchmarks}
\label{sec:bridges}

A \emph{Bridge} is a benchmark's direct implementation of \texttt{ActionableEnv}, and it is the \emph{only} layer in the system aware of the underlying runtime. We implement seven Bridges including four distinct runtime classes: a pure in-memory arithmetic game (Toy24), a text-adventure engine (ALFWorld via TextWorld), per-instance Docker containers (SWE-bench, OfficeQA and spreadsheetBench, where each \texttt{step} is a stateless \texttt{docker exec} against the task repository), and a Playwright-driven browser (WebArena via BrowserGym and Webshop). Everything above the Bridge, including the policy loop, the orchestrator, and all component code, is shared verbatim across the seven environments.

Each Bridge declares its action space as a \texttt{tool\_registry} of typed tools whose signatures are introspected into function-calling schemas for the policy prompt. The registry serves two roles that the design decouples: schema generation is universal, while \emph{dispatch} through it is optional. Toy24 routes \texttt{step()} through the registry because its state is the runtime. In contrast, ALFWorld, SWE-bench, and WebArena bypass it and drive their engine handle directly, since a TextWorld engine or a browser session cannot be threaded through the data-only state view. Bridges likewise choose their own persistence granularity along the two patterns above (full snapshot for Toy24, and reset-arguments-only for the three heavy runtimes) and publish a human-readable \texttt{env\_state\_schema()} describing the fields that component hooks may read. This schema is injected into the designer agent's prompt, closing the loop between what a Bridge exposes and what generated code can rely on.

\subsection{\texttt{EnvHarness}: Components as Composable Decorators}
\label{sec:envharness}

Each environment carries one \method, an ordered stack of components over its Bridge (Figure~\ref{fig:uml}, instance view). In code the stack is realized by the decorator pattern. \texttt{EnvHarness} is the abstract base every component derives from, and an \texttt{EnvHarness} \emph{is} an \texttt{ActionableEnv} that \emph{wraps} another \texttt{ActionableEnv}. Its default implementation delegates every interface method to the inner environment, so a concrete component overrides only the methods on the axes it affects. Because the wrapped object may itself carry components, they stack arbitrarily, meaning $\texttt{Rules}(\texttt{Setups}(\texttt{Toy24Bridge}))$ is again an \texttt{ActionableEnv}, and the policy interacting with the outermost layer cannot observe how many components sit beneath it. Persistence is layered accordingly: each component serializes only its own state, and a checkpoint records the environment plus an ordered list of components (innermost first), which the loader rebuilds inward-out by passing each reconstructed component as the \texttt{inner} field of the next. The three component types of Section~\ref{sec:components} ship as the classes \texttt{Setups}, \texttt{Rules}, and \texttt{Link} (Figure~\ref{fig:uml}, bottom right).

\paragraph{Setups: initial-state mutation by action replay.}
\texttt{Setups} realizes the $S_0$ axis of Eq.~\eqref{eq:setup} without any privileged access to environment internals. It carries the action list $\delta$. On \texttt{reset()}, it first resets the inner environment, then replays each action of $\delta$ through the ordinary \texttt{inner.step()} interface, and returns the post-replay observation as the episode's initial observation. The mutated start state is thus always a \emph{reachable} state, expressed in the environment's own action vocabulary rather than in a benchmark-specific state schema, so the saved form is nothing more than the action list itself. After replay, \texttt{Setups} invokes \texttt{notify\_replay\_complete()} so the inner environment can rewind per-episode counters (step budgets, repetition guards) that must not be charged for the preparation phase. Replay determinism is inherited from the seeded reset, ensuring the same \texttt{Setups} component reproduces the exact same start state across rollouts.

\paragraph{Rules: per-step I/O transformation hooks.}
\texttt{Rules} realizes the triple $(f_A, f_T, f_O)$ of Eq.~\eqref{eq:rule} as three pure-function hooks interposed on the step loop. \texttt{filter\_action(action, env\_state)} implements $f_A$ and may rewrite the agent's action or return a typed \texttt{Blocked} result before it reaches the inner environment. \texttt{modify\_transition(action, response, env\_state)} implements $f_T$ and rewrites the inner \texttt{EnvResponse}, and \texttt{filter\_observation(obs, env\_state)} implements $f_O$, transforming what the agent ultimately sees, including the initial observation at reset. All defaults are identities. A useful \texttt{Rules} component is a \emph{subclass} overriding some hooks, and this subclass is exactly what the designer agent emits as Python source. The saved state of a \texttt{Rules} component is therefore the source string itself. Loading recompiles it in a namespace that exposes only the abstract data types, and the compiled code executes inside a per-episode subprocess so that faulty generated code crashes an episode rather than the framework. Two boundaries are deliberate: blocked actions leave the environment untouched and return the current (re-observed, then $f_O$-filtered) state alongside the block reason, so a rejection never strands the policy. Furthermore, \texttt{Rules} does not implement $S_0$ or $R$, as initial state belongs to \texttt{Setups} and terminal task success remains the benchmark's own decision.

\paragraph{Link: composition for long-horizon episodes.}
Whereas \texttt{Setups} and \texttt{Rules} reshape a single task, \texttt{Link} composes \emph{two} \texttt{ActionableEnv}s into one episode, instantiating the composition logic $g$ of Eq.~\eqref{eq:link}. The handoff is decided by a per-step hook that either leaves the agent in the current sub-environment or routes it to another one, so serial concatenation, outcome-conditioned branching, and mid-task switching are all expressible; we use serial composition throughout this work. The agent interacts with environment $A$ until $A$'s task concludes, is then handed a spliced transition observation, and continues in environment $B$ under a shared step budget. \texttt{Link} masks the sub-environments' termination signals so that only the composite decides when the episode ends. It resets $B$ lazily at the handoff point, avoiding container or browser start-up cost when $A$ fails early, and caches each leg's outcome at its boundary so evaluation never re-runs an expensive scorer. The composite verdict is the conjunction $R' = R_A \wedge R_B$, each factor decided by the corresponding sub-environment's own verifier, so \texttt{evaluate()} succeeds only if both sub-tasks succeed and the chained task inherits trusted verification from both of its parts. Because \texttt{Link} calls nothing beyond the \texttt{ActionableEnv} contract (it never imports a concrete Bridge or inspects benchmark-specific fields), any pair of registered environments can be linked, including cross-benchmark pairs, turning single-task corpora into long-horizon trajectories with mid-episode task reorientation.

\section{Concrete Implementation Examples of the Chain (Link) Operator}
\label{app:link_examples}

To demonstrate the flexibility and programmability of the Chain operator (referred to as \texttt{Link} in our codebase), we present simplified Python implementations of different composition modes. Every mode is implemented by overriding the \texttt{modify\_transition} hook, which intercepts transitions after every step to decide whether to stay in the current environment or transition to a destination environment via \texttt{self.switch\_to()}.

\paragraph{Sequential Concatenation.}
By default, the \texttt{Link} operator concatenates environments in sequence. It automatically switches to the destination environment as soon as the first environment terminates, requiring no manual override of the transition logic.

\begin{componentbox}{Sequential Concatenation via Default Handoff}
# No custom transition override is needed. 
# The default Link hands off to EnvB automatically once EnvA terminates.
link = Link(EnvA(), EnvB(), a\_done\_via="terminated")
\end{componentbox}

\paragraph{Branching on Task Outcome.}
The composition logic can evaluate the success or failure of the first task upon termination, and dynamically route the agent to a harder task (such as \texttt{AdvancedEnv}) or an easier one (such as \texttt{RemedialEnv}).

\begin{componentbox}{Dynamic Branching on Outcome ($g$, routes based on success)}
class BranchOnOutcome(Link):
    def modify\_transition(self, action, response, env\_state):
        if not self.\_a\_is\_finished(response):
            return response  # Keep running the current task
        
        # Route to different environments based on the task outcome
        solved = self.env\_a.evaluate().success
        dest = AdvancedEnv() if solved else RemedialEnv()
        return self.switch\_to(dest)
\end{componentbox}

\paragraph{Dynamic Switch Mid-Task.}
The transition point is fully controlled by the harness. This allows switching the agent to a different environment immediately when a specific condition is met, without waiting for the current task to officially end.

\begin{componentbox}{Dynamic Switch Mid-Task ($g$, routes immediately on action)}
class SwitchOnAction(Link):
    def modify\_transition(self, action, response, env\_state):
        # Trigger an immediate switch when a specific action is observed
        if action.name == "trigger\_advanced\_mode":
            return self.switch\_to(AdvancedEnv())
        return response
\end{componentbox}

\paragraph{Interleaving Environments.}
Because the transition check runs after every single interaction step, the agent can alternate back and forth between two environments continuously during execution.

\begin{componentbox}{Interleaving Alternation ($g$, alternates every step)}
class Alternate(Link):
    def modify\_transition(self, action, response, env\_state):
        # Swap between Red and Blue environments on every single step
        next\_env = RedEnv() if isinstance(self.current\_env, BlueEnv) else BlueEnv()
        return self.switch\_to(next\_env)
\end{componentbox}

\section{Experiment Details}
\subsection{Benchmark Splits}
\label{app:splits}
Table~\ref{tab:splits} lists the training and evaluation splits. Training tasks are the corpus the designer agent reshapes; evaluation uses only original, unreshaped tasks.

\begin{table}[h]
\centering
\caption{Training and evaluation splits per benchmark.}
\label{tab:splits}
\begin{tabular}{l l l}
\toprule
\textbf{Benchmark} & \textbf{Training} & \textbf{Evaluation} \\
\midrule
ALFWorld            & 100 tasks from the standard train set & all remaining held-out tasks \\
WebArena            & 20 tasks per sub-domain               & all remaining tasks \\
SWE-bench           & 100 tasks from SWE-bench Lite         & 407 Verified issues not in Lite \\
OfficeQA            & 50 tasks (official split)             & 172 official test tasks \\
SpreadsheetBench    & 100 of the 400 verified tasks         & 299 held-out tasks (897 instances) \\
\bottomrule
\end{tabular}
\end{table}

For SpreadsheetBench, Pass@1 aggregates over base tasks and Mean Score averages over all instances. On ALFWorld, In-Dist and OOD refer to the benchmark's own \textit{seen} and \textit{unseen} evaluation splits, which differ in whether the task's object--receptacle configuration appeared during training; we do not construct these splits ourselves. Skill extraction uses the same model as the designer and the policy. 

\subsection{Baseline Details}
\label{app:baselines}
GenEnv~\citep{guo2025genenv} generates new tasks with an environment-simulator model that keeps task difficulty at the edge of the agent's current ability. VeriEnv~\citep{chae2026safe} clones websites into executable synthetic environments whose rewards are checked programmatically. SWE-smith~\citep{yang2026swe} synthesizes new repository-level task instances. We run each with the same seed tasks and the same model as \method, and each produces the same number of environments as \method does, so differences come from the generation strategy rather than the data, the model, or the amount of generation. Generation pipelines are benchmark-specific because they must reach into an environment's internals and construct verifiers, which is why each baseline covers a single benchmark and none exists for the office domain. For every baseline, skill extraction, retrieval, and the policy model are identical to ours, and only the environment the skills come from differs.
\subsection{\designer{} Hyperparameters}
\label{app:rigger_hparams}

Table~\ref{tab:rigger_hparams} collects the settings of the \designer{}. The same values are used on every benchmark.

\begin{table}[h]
\centering
\caption{\designer{} hyperparameters. The same settings are used across all benchmarks.}
\label{tab:rigger_hparams}
\begin{tabular}{lll}
\toprule
\textbf{Stage} & \textbf{Parameter} & \textbf{Value} \\
\midrule
Observe  & Baseline rollouts per task ($K$)         & 5 \\
\midrule
Write    & Components per candidate                 & unbounded (designer's choice) \\
\midrule
Validate & Fresh rollouts per candidate ($K$)       & 5 \\
         & Revision budget (write--validate rounds) & 5 \\
\midrule
General  & Designer backbone                        & same as policy \\
\bottomrule
\end{tabular}
\end{table}

\paragraph{Observe.} Before its first proposal the designer sees $K=5$ rollouts of $\pi$ on the unmodified task, together with the resulting success rate and per-rollout outcomes. This baseline serves three purposes made explicit in the prompt: it establishes whether the policy can solve the task at all, how much headroom its current solution leaves, and which parts of the environment it actually exercises.

\paragraph{Write.} A candidate is a set of one or more components emitted together; we place no cap on the set size, and the designer decides how many components a diagnosis calls for. A candidate is accepted or rejected as a whole rather than component by component.

\paragraph{Validate.} Each candidate is evaluated on a fresh set of $K=5$ rollouts under the same settings, so its success rate is directly comparable to the baseline. Acceptance is decided from these $K$ trajectories in aggregate---success rate, failure distribution, and timeout count---never from a single trace. A candidate that is neither accepted nor rejected returns to the Write stage with the validation trajectories attached; this write-and-validate loop runs at most 5 times per instance, after which the instance yields no component.
\section{Analyses Details}
This appendix collects the supplementary material for the analyses of Section~\ref{sec:analysis}: the protocol and full results behind the reported numbers, the additional experiments referred to there, and representative skills extracted in each setting.

\subsection{Experimental Details for Reinforcement Learning}
\label{sec:appendix_rl_details}

We evaluated the effectiveness of \method environments in a Reinforcement Learning (RL) setting. We utilized Group Relative Policy Optimization (GRPO) to train a Qwen3-8B-base model on the ALFWorld and Webshop benchmark. The specific hardware configurations and training hyperparameters are detailed below:

\paragraph{Hardware and System Configuration.}
The RL experiments were conducted on a single compute node equipped with 8$\times$ NVIDIA H100 GPUs. For rollout generation, we utilized the vLLM framework with Tensor Parallelism (TP) set to 1, GPU memory utilization configured to 0.5, and eager execution enforced. To optimize memory usage during training, we enabled Fully Sharded Data Parallel (FSDP) alongside parameter offloading, optimizer offloading, and gradient checkpointing.

\paragraph{Hyperparameters.}
\begin{itemize}[leftmargin=*, itemsep=2pt, parsep=0pt]
    \item \textbf{Batch Size:} The global training batch size was set to 16, with a PPO mini-batch size of 256. Both the PPO micro-batch size and the log-probability micro-batch size were set to 4 per GPU.
    \item \textbf{Sequence Length:} The maximum prompt length was constrained to 4096 tokens, and the maximum response length was set to 512 tokens.
    \item \textbf{Environment Settings:} We used the \method-integrated  environment, configured with a history length of 50 and a maximum episode length of 50 steps. The random seed was fixed at 0.
    \item \textbf{Reward and Sampling:} An invalid action penalty with a coefficient of 0.1 was applied to discourage unexecutable actions. Rollout sampling was performed with a temperature of 0.4.
    \item \textbf{Training Schedule:} The model was trained for a total of 150 epochs (equivalent to 150 steps, as one epoch corresponds to one step based on the training batch configuration).
\end{itemize}

\subsection{Skills Across Co-evolution Rounds}
\label{app:coevo_skills}
Each round of the loop diagnoses flaws the previous policy did not yet show, so the skills extracted per round shift in focus, from basic test invocation and file editing, to keeping the test loop alive when the runner itself breaks, to interpreter resolution and pre-edit navigation. The gains shrink round over round (Figure~\ref{fig:coevo}) as the remaining flaws grow more local. Every skill below traces to a specific accepted component. The banks also contain skills distilled from rollouts that no component targeted, and we show the component-driven ones here so the causal link between a written component and the induced skill is inspectable. Each component's code follows its skill in an amber box, together with the axis of Eq.~\eqref{eq:rule} it exercises. Import lines are omitted from the listings.

\paragraph{Round 1.}
Distilled from the base policy's first-pass failures. The policy does not yet invoke tests reliably or apply file edits within the available tools, and the round-1 components shape exactly those two surfaces.
\begin{skillbox}[pairtop]{Use \texttt{pytest -x} to prevent test suite timeouts}
\skillfield{Description} When running a large or potentially hanging test suite, use the exit-on-first-failure flag to get immediate feedback and avoid environment timeouts. \\
\skillfield{Content} When running tests that may hang or take too long, run \texttt{pytest -x} (or \texttt{pytest --exitfirst}) to stop execution instantly on the first failing test.
\skillorigin{Forced by the component below. It rewrites every unguarded pytest command to include \texttt{-x}, and the extracted skill is the very pattern the component enforces.}
\end{skillbox}
\begin{componentbox}{The accepted component behind it ($f_A$, enforces fail-fast test runs)}
class _Rules(Rules):
    def filter_action(self, action, env_state):
        if action.name == "bash":
            command = action.kwargs.get("command", "")
            if ("pytest" in command
                    and "-x" not in command
                    and "--maxfail" not in command):
                # Add -x to fail fast and prevent 60s docker exec
                # timeouts on compatibility hangs.
                action.kwargs["command"] = command.replace(
                    "pytest", "pytest -x")
        return action
\end{componentbox}
\begin{skillbox}[pairtop]{Use \texttt{patch -p1 << 'EOF'} for precise file modifications}
\skillfield{Description} When structured editing tools are unavailable, apply code changes and test additions reliably using standard patch inputs. \\
\skillfield{Content} Use \texttt{patch -p1 << 'EOF'} followed by a standard unified diff to apply precise multi-line modifications to source and test files.
\skillorigin{Forced by the component below. It closes the three edit paths the base policy reaches for, and the patch workflow is the remaining escape.}
\end{skillbox}
\begin{componentbox}{The accepted component behind it ($f_A$, blocks the default editing paths)}
class _Rules(Rules):
    def filter_action(self, action, env_state):
        if action.name == "bash":
            cmd = action.kwargs.get("command", "")

            # 1. Inline python file editing via `python -c` is off-limits.
            if "python" in cmd and "-c" in cmd:
                if "open" in cmd and ("write" in cmd or "replace" in cmd):
                    return Blocked(reason=(
                        "Inline python file editing (-c) is restricted "
                        "to prevent file corruption. Use standard file "
                        "utilities or patch/diff tools."))

            # 2. `conda run` disabled; invoke env binaries by absolute path.
            if "conda run" in cmd:
                return Blocked(reason=(
                    "'conda run' is disabled in this environment due to "
                    "subshell execution constraints. Invoke the "
                    "environment's binaries directly using their "
                    "absolute paths."))

            # 3. Direct execution of custom Python scripts is disabled
            #    (except when driving pytest / unittest).
            if re.search(r"\bpython[0-9.]*\s+[^-\s][^\s]*\.py\b", cmd):
                if "pytest" not in cmd and "unittest" not in cmd:
                    return Blocked(reason=(
                        "Direct execution of custom Python script files "
                        "(e.g. 'python script.py') is disabled to keep "
                        "standard environment execution. Use standard "
                        "command-line tools (patch, sed, redirected cat) "
                        "to modify files, and run tests via pytest."))
        return action
\end{componentbox}
\begin{skillbox}[pairtop]{Target specific test files to avoid execution timeouts}
\skillfield{Description} When running tests in a large codebase, specify the exact test file path instead of the parent directory to prevent execution timeouts. \\
\skillfield{Content} Run tests targeting the exact file, e.g., \texttt{pytest path/to/test\_file.py} or \texttt{python -m unittest path/to/test\_file.py}, instead of running the entire directory.
\skillorigin{Forced by the component below. It kills every whole-suite invocation with a simulated timeout, and per-file targeting is the escape.}
\end{skillbox}
\begin{componentbox}{The accepted component behind it ($f_T$, kills broad-scope test runs)}
class _Rules(Rules):
    def modify_transition(self, action, raw_response, env_state):
        if action.name != "bash":
            return raw_response
        cmd = action.kwargs.get("command", "")

        # Broad-scope test invocations get killed; force per-file
        # targeting of the relevant test module.
        if (("bin/test" in cmd or "sympy.test" in cmd)
                and "test_polysys" not in cmd):
            return EnvResponse(
                observation=Observation(
                    text="[docker exec timed out after 60s]\n",
                    data=raw_response.observation.data),
                reward=raw_response.reward,
                terminated=raw_response.terminated,
                truncated=raw_response.truncated,
                info={**raw_response.info, "last_returncode": 124},
            )
        return raw_response
\end{componentbox}

\paragraph{Round 2.}
The round-1 policy runs targeted, fail-fast tests and applies diffs through patch, so those failures largely disappear. The residual failures sit one level up. The test entrypoint itself may break, even a targeted run can be killed by the resource limiter, and the file argument itself is policed. Each component below imposes one of these constraints, and each skill is the escape the policy found. None of these failure modes appear in the round-1 bank, which assumed a working command line.
\begin{skillbox}[pairtop]{Invoke pytest programmatically via \texttt{pytest.main} when the CLI is broken}
\skillfield{Description} When the standard \texttt{pytest} command-line entrypoint is broken, missing, or misconfigured, run tests programmatically through the interpreter. \\
\skillfield{Content} Use \texttt{python -c "import pytest; pytest.main(['<test\_file>'])"} to execute specific test suites directly when the \texttt{pytest} executable fails to run.
\skillorigin{Forced by the component below. It breaks the pytest entrypoint, and this workaround is the only path it leaves open.}
\end{skillbox}
\begin{componentbox}{The accepted component behind it ($f_T$, breaks the pytest entrypoint)}
class _Rules(Rules):
    def modify_transition(self, action, raw_response, env_state):
        if action.name != "bash":
            return raw_response
        cmd = action.kwargs.get("command", "")

        # Only intercept "raw" pytest invocations; leave the `python -c`
        # workaround untouched.
        if "pytest" not in cmd or "python -c" in cmd or "patch" in cmd:
            return raw_response

        # `python -m pytest ...` -> module missing.
        if re.search(r"\bpython(3)?\s+-m\s+pytest\b", cmd):
            stdout = "No module named pytest\n"
            return EnvResponse(
                observation=Observation(
                    text=stdout, data=raw_response.observation.data),
                reward=raw_response.reward,
                terminated=raw_response.terminated,
                truncated=raw_response.truncated,
                info={**raw_response.info, "exit_code": 1,
                      "result": {"stdout": stdout, "exit_code": 1}},
            )

        # Bare `pytest ...` -> command not found.
        if re.search(r"\bpytest\b", cmd):
            stdout = "bash: pytest: command not found\n"
            return EnvResponse(
                observation=Observation(
                    text=stdout, data=raw_response.observation.data),
                reward=raw_response.reward,
                terminated=raw_response.terminated,
                truncated=raw_response.truncated,
                info={**raw_response.info, "exit_code": 127,
                      "result": {"stdout": stdout, "exit_code": 127}},
            )

        return raw_response
\end{componentbox}
\begin{skillbox}[pairtop]{Use \texttt{pytest -k} to filter test cases and avoid process kills}
\skillfield{Description} When running an entire test file is killed (exit code 137) or times out under resource limits, run only the relevant test cases. \\
\skillfield{Content} Use \texttt{pytest <file> -k "pattern1 or pattern2"} to run specific test cases and avoid resource exhaustion.
\skillorigin{Forced by the component below. It simulates resource kills on whole-file runs, and the filter is the whitelisted escape.}
\end{skillbox}
\begin{componentbox}{The accepted component behind it ($f_T$, simulates resource kills)}
class _Rules(Rules):
    def modify_transition(self, action, raw_response, env_state):
        if action.name != "bash":
            return raw_response
        cmd = action.kwargs.get("command", "")
        if "pytest" not in cmd:
            return raw_response

        # Whole-file pytest runs (no -x / no -k) simulate an OOM kill.
        if "-x" not in cmd and "-k" not in cmd:
            new_info = {**raw_response.info}
            if isinstance(new_info.get("result"), dict):
                new_info["result"] = {
                    **new_info["result"],
                    "exit_code": 137,
                    "stdout": "",
                    "stderr": "Killed\n",
                }
            return EnvResponse(
                observation=Observation(
                    text="Killed\n", data=raw_response.observation.data),
                reward=raw_response.reward,
                terminated=raw_response.terminated,
                truncated=raw_response.truncated,
                info=new_info,
            )
        return raw_response
\end{componentbox}
\begin{skillbox}[pairtop]{Use nonexistent files in \texttt{pytest.main} to test CLI options}
\skillfield{Description} When testing pytest CLI option parsing or configuration loading programmatically, pass a nonexistent filename to \texttt{pytest.main()} to prevent expensive test collection. \\
\skillfield{Content} Use \texttt{python -c "import pytest; pytest.main(['--your-opt', 'nonexistent.py'])"} to verify CLI argument parsing and option registration without scanning the workspace.
\skillorigin{Forced by the component below. It times out any pytest run that names no file, and on this task, inside the pytest codebase itself, naming a nonexistent file verifies option registration without paying the collection cost.}
\end{skillbox}
\begin{componentbox}{The accepted component behind it ($f_T$, requires a file target)}
class _Rules(Rules):
    def modify_transition(self, action, raw_response, env_state):
        if action.name != "bash":
            return raw_response
        cmd = action.kwargs.get("command", "")

        # If they try to run pytest globally or on the whole test
        # directory without targeting a specific file, time out fast.
        markers = ["testing/", ".py", "-h", "--help"]
        if "pytest" in cmd and not any(m in cmd for m in markers):
            return EnvResponse(
                observation=Observation(
                    text=("Error: Command timed out (limit of 15 seconds "
                          "exceeded). Please target specific test files "
                          "to avoid timeouts."),
                    data=raw_response.observation.data),
                reward=raw_response.reward,
                terminated=raw_response.terminated,
                truncated=raw_response.truncated,
                info={**raw_response.info, "exit_code": 124},
            )
        return raw_response
\end{componentbox}

\textbf{Round 3.} The round-2 policy drives the test runner robustly. What remains sits below and around the shell, interpreter resolution through PATH and the navigation habits needed once cheap in-place edits are taken away. On about a third of the training tasks the round-2 policy now succeeds on every baseline rollout; \designer treats this as a signal to make the environment harder, but its candidates are rejected at validation, leaving the success rate untouched or driving it to zero, and the write-and-validate loop exhausts its revision budget without an accepted component. These tasks therefore contribute nothing to the round-3 bank.
\begin{skillbox}[pairtop]{Use the active conda environment's absolute binary path}
\skillfield{Description} When global \texttt{python} or \texttt{pip} commands fail due to version mismatches, locate and run the specific conda environment's binaries directly. \\
\skillfield{Content} Prepend the environment's bin directory to \texttt{PATH} or invoke it by absolute path, e.g., \texttt{export PATH=/opt/miniconda3/envs/<env\_name>/bin:\$PATH} or \texttt{/opt/miniconda3/envs/<env\_name>/bin/python}.
\skillorigin{Forced by the component below. It rewrites every command to resolve the interpreter through the environment's own bin directory, and the extracted skill is the pattern the component enforces. Earlier rounds assumed the shell already resolved \texttt{python} correctly; this skill is the first to reach below that assumption.}
\end{skillbox}
\begin{componentbox}{The accepted component behind it ($f_A$, pins interpreter resolution)}
class _Rules(Rules):
    def filter_action(self, action, env_state):
        if action.name == "bash" and "command" in action.kwargs:
            cmd = action.kwargs["command"]
            # Prepend the testbed environment's bin directory so that
            # `python` and `pytest` resolve to the right interpreter.
            if "/opt/miniconda3/envs/testbed/bin" not in cmd:
                action.kwargs["command"] = (
                    "export PATH=/opt/miniconda3/envs/testbed/bin:$PATH"
                    f" && {cmd}"
                )
        return action
\end{componentbox}
\begin{skillbox}[pairtop]{Trace exception propagation using \texttt{grep -rn} on handler definitions}
\skillfield{Description} When an exception is swallowed or improperly handled, trace its propagation path by searching for the exception handler or wrapping function definitions across the codebase. \\
\skillfield{Content} Use \texttt{grep -rn "def convert\_exception\_to\_response"} or \texttt{grep -rn "except <ExceptionName>"} to locate the exact middleware or wrapper handling the exception flow.
\skillorigin{Forced by the component below. It disables in-place \texttt{sed} edits, so every edit costs a full patch, and the policy learns to invest in precise navigation before committing to one.}
\end{skillbox}
\begin{componentbox}{The accepted component behind it ($f_A$, disables in-place sed edits)}
class _Rules(Rules):
    def filter_action(self, action, env_state):
        if action.name == "bash":
            cmd = action.kwargs.get("command", "")
            # Detect `sed -i` / `sed --in-place` (any spacing variant)
            # and swap it for a fake BSD-sed error, forcing the policy
            # onto safer editing paths.
            if "sed " in cmd and ("-i" in cmd or "--in-place" in cmd):
                action.kwargs["command"] = (
                    "echo 'sed: illegal option -- i' && exit 1"
                )
        return action
\end{componentbox}
\begin{skillbox}{Locate reference implementations using \texttt{grep -rn} on registration functions}
\skillfield{Description} When implementing or testing a framework-specific component, search the test suite for existing registrations of that component to find working examples. \\
\skillfield{Content} Use \texttt{grep -rn "register\_converter" tests/} (or similar registration and decorator patterns) to quickly find reference implementations and test cases for framework components.
\skillorigin{Forced by the same component shown above. The sed block pushed the policy into grep-based investigation, and this skill pursues a different goal, finding working examples to imitate before writing new framework glue.}
\end{skillbox}

\subsection{Skills from \textit{Chain} Environments}
\label{app:link_skills}
Skills extracted from \textit{Chain} environments concern behaviors that only appear when tasks are joined, such as managing a shared step budget and reorienting after a mid-episode task switch. We list representative examples below.

\begin{skillbox}{Manage shared step budgets across joined tasks}
\skillfield{Description} Treat the joined task structure as a single, finite budget, prioritizing ``good enough'' solutions in the first task to ensure sufficient resources for the second. \\
\skillfield{Content} The agent successfully completed two distinct tasks within a single episode. By efficiently resolving the first task (fixing the \texttt{\_makepath} issue in \texttt{pytest}), the agent preserved enough steps to handle the significant environment-setup challenges (dependency issues, circular imports, and missing attributes) encountered in the second task (\texttt{RidgeClassifierCV} in \texttt{scikit-learn}). This demonstrates the importance of maintaining momentum and not over-optimizing the first task at the expense of the second.
\end{skillbox}

\begin{skillbox}{Re-orient environment after task handoff}
\skillfield{Description} Immediately perform environment reconnaissance (e.g., \texttt{conda env list}, \texttt{which python}) upon receiving a new task to identify the correct test runner and environment configuration, as these often differ between repositories. \\
\skillfield{Content} When the agent transitions to a new task (e.g., from \texttt{matplotlib} to \texttt{django}), it cannot assume the previous environment's \texttt{pytest} or \texttt{python} paths are valid. In this trajectory, the agent correctly identified that the \texttt{django} repository required a specific \texttt{runtests.py} script and a different conda environment path, avoiding the ``command not found'' errors that occurred when it initially tried to reuse the \texttt{matplotlib} test-running conventions.
\end{skillbox}

\subsection{Cross-Model Results}
\label{app:crossmodel}
Table~\ref{tab:crossmodel} lists the success rate and the average episode length for all four policy models under the protocol of Section~\ref{sec:analysis}. The success rates move in one direction, while the average steps reveal three different regimes. Qwen3.6 27B runs 69.8 steps bare, the longest of any model, and skills nearly halve this to 37.1, so the bare policy spends most of its budget on undirected trial and error that the skills replace with known procedures. Gemini 3.1 Flash-Lite shows the reverse. The bare policy gives up early at 36.7 steps, and skills make it persist at around 50 steps while solving far more tasks, so here the extra length is the point. Claude Sonnet 4.6 barely moves, from 29.3 to about 25 steps, since a policy that is already directed has little dead time for skills to reclaim. Gemini 3.5 Flash sits between these regimes and is the one model where \method skills improve both metrics at once against both baselines, more resolved issues in visibly shorter episodes.

Two further points follow from the same numbers. First, the gains of \method over the original-environment skills do not come from simply running longer. The episode lengths of the two skill sources are within a step of each other on Flash-Lite and Sonnet, \method is over five steps shorter on Flash, and only on Qwen does it spend more, 3.7 extra steps for 3.7 extra points. Across models the additional successes arrive within essentially the same budget. Second, average steps alone is not a quality signal. A short episode can mean an efficient solution, as for Sonnet, or an early surrender, as for bare Flash-Lite, and the two cases sit at nearly the same step count. Reading the metric therefore requires the success rate next to it, which is why we report both here rather than in the main text. In short, skills shorten episodes where the policy flails, lengthen them where it quits, and leave them alone where the policy already knows where it is going.

\begin{table}[h]
\centering
\caption{Cross-model results on SWE-bench Verified under the same protocol as Table~\ref{tab:main_results_swe}; the Gemini 3.5 Flash column reports the same runs as that table, rounded to one decimal. SR is success rate and AS is average steps. Best in \textbf{bold}.}
\label{tab:crossmodel}
\resizebox{\textwidth}{!}{
\begin{tabular}{l cc c cc c cc c cc}
\toprule
\multirow{2}{*}{\textbf{Skill Source}} & \multicolumn{2}{c}{\textbf{Gemini 3.1 Flash-Lite}} & & \multicolumn{2}{c}{\textbf{Qwen3.6 27B}} & & \multicolumn{2}{c}{\textbf{Gemini 3.5 Flash}} & & \multicolumn{2}{c}{\textbf{Claude Sonnet 4.6}} \\
\cmidrule{2-3} \cmidrule{5-6} \cmidrule{8-9} \cmidrule{11-12}
& \textbf{SR ($\uparrow$)} & \textbf{AS ($\downarrow$)} & & \textbf{SR ($\uparrow$)} & \textbf{AS ($\downarrow$)} & & \textbf{SR ($\uparrow$)} & \textbf{AS ($\downarrow$)} & & \textbf{SR ($\uparrow$)} & \textbf{AS ($\downarrow$)} \\
\midrule
No Skills      & 30.7 & \textbf{36.7} & & 41.0 & 69.8 & & 47.7 & 53.6 & & 67.2 & 29.3 \\
Original Envs  & 36.8 & 50.0 & & 48.4 & \textbf{37.1} & & 49.9 & 55.0 & & 69.2 & \textbf{25.4} \\
\method{} Envs & \textbf{40.0} & 50.6 & & \textbf{52.1} & 40.8 & & \textbf{52.6} & \textbf{49.6} & & \textbf{72.4} & 25.6 \\
\bottomrule
\end{tabular}
}
\end{table}
\newpage
\section{Additional Analysis}
\label{app:add_analysis}
\paragraph{\method Produces More Generalizable Skills.}
\begin{wraptable}{r}{0.6\textwidth}
\vspace{-4mm}
\centering
\small
\setlength{\tabcolsep}{3.5pt}
\begin{tabular}{llccc}
\toprule
\textbf{Benchmark} & \textbf{Held-out Type} & \textbf{Orig.} & \textbf{\method} & \textbf{$\Delta$} \\
\midrule
\multirow{7}{*}{ALFWorld}
& clean      & 54.8 & \textbf{71.2} & +16.4 \\
& cool       & 38.5 & \textbf{39.3} & +0.8 \\
& heat       & \textbf{61.1} & 52.4 & -8.7 \\
& look\_lamp  & 79.0 & \textbf{82.7} & +3.7 \\
& simple     & \textbf{83.6} & \textbf{83.6} & 0.0 \\
& two\_obj    & 46.4 & \textbf{52.9} & +6.5 \\
\cmidrule{2-5}
& \textit{Average} & 60.6 & \textbf{63.7} & \textbf{+3.1} \\
\bottomrule
\end{tabular}
\vspace{1mm}
\caption{Leave-one-out generalization on ALFWorld. Skills are extracted from environments of all task types except the held-out type, then evaluated on it.}
\label{tab:held_out}
\vspace{-2mm}
\end{wraptable}
To test whether the skills transfer beyond the task types they were learned on, we run a leave-one-out evaluation on ALFWorld. Skills are extracted from environments covering every task type except one, and the policy is evaluated on the held-out type alone, so any gain must come from behaviors that carry across types rather than from familiarity with the held-out tasks. As Table~\ref{tab:held_out} shows, skills from \method environments outperform skills from the original environments on four of the six types and by 3.1 points on average, with the largest gain of 16.4 points on \textit{clean} and one regression of 8.7 points on \textit{heat}. Reshaped environments push the policy off its memorized routines during extraction, so the resulting skills encode behaviors that apply across task types instead of recipes tied to a single one.

\paragraph{\method{} Maintains Practical Compute Overhead.}
\begin{wraptable}{r}{0.7\textwidth}
\centering
\small
\setlength{\tabcolsep}{4pt}
\renewcommand{\arraystretch}{1.1}
\begin{tabular}{llrrr}
\toprule
\textbf{Benchmark} & \textbf{Method} & \textbf{Design Tok.} & \textbf{Rollout Tok.} & \textbf{Total Tok.} \\
\midrule
\multirow{2}{*}{ALFWorld}
& GenEnv & 38K & 64.2M & 64.2M \\
& \method{} & 1.46M & 226.6M & 228.0M \\
\midrule
\multirow{2}{*}{WebArena}
& VeriEnv & 20K & 137.7M & 137.8M \\
& \method{} & 1.58M & 135.7M & 137.3M \\
\bottomrule
\end{tabular}
\vspace{1mm}
\caption{Estimated token consumption. Design tokens cover the calls that propose and refine components; rollout tokens cover every interaction a method drives. These rollouts are not the same kind of work across rows: \method{} and VeriEnv execute them against the real environment, whereas GenEnv's are LLM-simulated.}
\label{tab:compute_overhead_baselines}
\vspace{-2mm}
\end{wraptable}
Table~\ref{tab:compute_overhead_baselines} decomposes token consumption into \textit{Design Tokens} and \textit{Rollout Tokens}. \method{} spends far more on design than the single-pass baselines (1.46M vs.\ 38K on ALFWorld), a deliberate cost of feeding full trajectories into its prompts to diagnose weaknesses rather than generating tasks blind. Design is nonetheless a small share of either budget, and rollouts dominate.

Against VeriEnv, which also executes in the real environment, the totals are essentially the same (137.3M vs.\ 137.8M), so the gains of Section~\ref{sec:experiments} do not come from outspending the baseline. GenEnv's total is 3.5$\times$ lower, but its rollouts are simulated rather than executed, and that saving buys the hallucinated transitions and drifting success signals discussed in Appendix. The extra cost is thus the cost of grounding: at equal grounding \method{} matches the baseline's footprint, and where it spends more, it spends on real execution against a trusted verifier.
\newpage

\paragraph{\method Reshapes Environments Toward Objective Metric Targets.}
\begin{wraptable}{r}{0.55\textwidth}
\centering
\small
\setlength{\tabcolsep}{4pt}
\begin{tabular}{llcc}
\toprule
\textbf{Metric} & \textbf{Band} & \textbf{Orig.} & \textbf{\method} \\
\midrule
Success rate (SR) & $[0.4, 0.6]$ & 6.0 & \textbf{80.0} \\
Avg.\ steps (AS)  & $[25, 35]$   & 18.0 & \textbf{53.0} \\
\bottomrule
\end{tabular}
\vspace{1mm}
\caption{Objective metric targeting on ALFWorld. Entries are the percentage of tasks whose measured value falls inside the target band, before and after reshaping.}
\label{tab:metric_target}
\vspace{-2mm}
\end{wraptable}
Beyond skill quality, we ask whether \method can steer an environment so that an objective, quantitative metric lands in a prescribed range. We run the loop on 100 ALFWorld tasks under two such metrics, measuring each task with $K=10$ rollouts: per-task success rate (SR), targeted to $[0.4, 0.6]$, and the average number of steps on successful episodes (AS), targeted to $[25, 35]$. In both cases \method reshapes a task in whichever direction its baseline value requires, tightening it when the value sits above the band and scaffolding it when the value sits below. Table~\ref{tab:metric_target} reports the fraction of tasks landing inside the band. SR is the more tractable target: the original tasks are strongly bimodal, with most either always solved or never solved, and \method compresses them into the middle of the range, raising in-band coverage from 6.0\% to 80.0\% and moving mean SR from 0.74 to 0.48. AS is a tighter constraint, since it fixes an exact step count rather than a rate, yet coverage still rises from 18.0\% to 53.0\%. A single interface therefore suffices to calibrate an environment against an explicit, measurable objective, without any access to its internals.

\paragraph{Teaching against a specified weakness.}
\label{app:on_demand}
For each case we hand the designer one sentence naming a capability weakness. The designer writes components that make the weakness fatal inside an ordinary benchmark task, runs the policy in the reshaped environment, and one skill is distilled from the resulting trajectories. Table~\ref{tab:on_demand_targeted} gives an overview, and the rest of this appendix lists the remaining cases in full, the generated component next to the skill it produced. The SWE-bench verification case appears in the main text. The designer chose its own component axes without instruction, staging start states with a Stage, blocking shortcuts with the action filter of a Contract, and faking consequences with the transition hook of a Contract. Every component stays inside the source distribution and leaves goals and scorers untouched. Code is abridged for space. The policy and the designer are Gemini 3.1 Flash-Lite on ALFWorld and WebArena and Gemini 3.5 Flash on SWE-bench Verified.

\begin{table}[h]
\centering
\caption{Nine specified weaknesses, the component the designer generated, and the skill distilled from the trajectories collected in the reshaped environment.}
\label{tab:on_demand_targeted}
\small
\setlength{\tabcolsep}{4pt}
\renewcommand{\arraystretch}{1.25}
\begin{tabular}{@{}l p{4.0cm} p{4.2cm} p{3.7cm}@{}}
\toprule
\textbf{Axis} & \textbf{Specified weakness} & \textbf{Generated component} & \textbf{Distilled skill} \\
\midrule
\multicolumn{4}{@{}l}{\cellcolor{hdr}\textbf{ALFWorld}} \\
\addlinespace[2pt]
Stage & Takes objects from closed containers without opening them
& Target object starts inside a closed drawer
& Pre-Interaction State Verification \\
\addlinespace[3pt]
Stage & Searches containers in an inefficient order
& Three drawers pre-opened to stage an ordering
& Semantic Container Prioritization \\
\addlinespace[3pt]
Stage & Forgets the second object in multi-object tasks
& First sub-goal completed in advance
& Task-State Verification Loop \\
\midrule
\multicolumn{4}{@{}l}{\cellcolor{hdr}\textbf{WebArena}} \\
\addlinespace[2pt]
Contract $f_A$ & Concludes without scrolling to content below the fold
& Retrieval actions blocked until a scroll happens
& Incremental Viewport Expansion \\
\addlinespace[3pt]
Stage & Counts paginated rows by hand instead of filtering
& Episode starts on the order grid, filter bar in view
& Query-Based Data Filtering \\
\addlinespace[3pt]
Contract $f_A$ & Guesses URLs instead of using the site search
& Direct navigation blocked
& Search-First Navigation Protocol \\
\midrule
\multicolumn{4}{@{}l}{\cellcolor{hdr}\textbf{SWE-bench Verified}} \\
\addlinespace[2pt]
Stage, $f_A$ & Edits the wrong function without reading test fixtures
& Test file restructured, git resets blocked
& Context-Aware Code Modification \\
\addlinespace[3pt]
Contract $f_T$ & Submits a patch without running the failing test
& Submission rejected until the tests have run
& Verification-Driven Development Loop \\
\addlinespace[3pt]
Contract $f_T$ & Uses \texttt{sed -i} and corrupts indentation
& File silently corrupted when \texttt{sed} is used
& Safe File Modification via Python Scripting \\
\bottomrule
\end{tabular}
\end{table}

\paragraph{ALFWorld.}

\begin{weaknessbox}
The policy takes objects from closed containers without opening them first, wasting turns.
\end{weaknessbox}
\begin{componentbox}{Generated component (Stage, $\delta$ axis)}
# The target now starts inside a closed drawer, so the
# habit fails on the first attempt.
delta = ["go to drawer 1", "close drawer 1"]
\end{componentbox}
\begin{skillbox}{Pre-Interaction State Verification}
\skillfield{Description} When the agent intends to manipulate an object contained within or covered by another object, or when an interaction fails due to an obstruction. \\
\skillfield{Content} Before executing a take or manipulate command, perform an examine or open action on the target container to verify its state and ensure the object is accessible.
\end{skillbox}

\begin{weaknessbox}
The policy searches containers in an inefficient order, not prioritizing the locations most likely to hold the target.
\end{weaknessbox}
\begin{componentbox}{Generated component (Stage, $\delta$ axis)}
# Three drawers are pre-opened, staging an ordering in the
# initial observation.
delta = ["go to drawer 1", "open drawer 1",
         "go to drawer 2", "open drawer 2",
         "go to drawer 3", "open drawer 3"]
\end{componentbox}
\begin{skillbox}{Semantic Container Prioritization}
\skillfield{Description} When searching for multiple instances of an object type across a room with many potential storage locations. \\
\skillfield{Content} Prioritize visiting surfaces such as countertops and tables before closed containers such as drawers and cabinets, to maximize visibility and minimize the open and close interactions needed to locate all target items.
\end{skillbox}

\begin{weaknessbox}
The policy fails multi-object tasks. After placing the first object it forgets the second and ends early.
\end{weaknessbox}
\begin{componentbox}{Generated component (Stage, $\delta$ axis)}
# The first sub-goal is already done at episode start, so
# the task now hinges on remembering the second.
delta = ["go to countertop 1",
         "take potato 1 from countertop 1"]
\end{componentbox}
\begin{skillbox}{Task-State Verification Loop}
\skillfield{Description} When the agent completes a sub-goal in a multi-step task and needs to determine whether the overall objective is fully satisfied. \\
\skillfield{Content} After every completed sub-goal, re-examine the original task description and the current environment state to identify remaining unfulfilled requirements before ending the episode.
\end{skillbox}

\paragraph{WebArena.}

\begin{weaknessbox}
The policy concludes without scrolling, missing results below the fold.
\end{weaknessbox}
\begin{componentbox}{Generated component (Contract, $f_A$ axis)}
class _Rules(Rules):
    def filter_action(self, action, env_state):
        if is_scroll(action):
            env_state.extras["has_scrolled"] = True
            return action
        if not env_state.extras.get("has_scrolled") \
                and is_retrieval(action):  # click, fill, select
            return Blocked("Scroll down first so all "
                           "content is visible.")
        return action
\end{componentbox}
\begin{skillbox}{Incremental Viewport Expansion}
\skillfield{Description} When a task requires counting or extracting data from a list that may be paginated or truncated by the viewport. \\
\skillfield{Content} Execute scroll-to-bottom actions followed by DOM re-inspection to trigger lazy loading and reveal hidden elements before finalizing the extraction.
\end{skillbox}

\begin{weaknessbox}
The policy counts paginated order rows by hand instead of applying date and status filters, losing track across pages.
\end{weaknessbox}
\begin{componentbox}{Generated component (Stage, $\delta$ axis)}
# The episode starts on the order grid, with its filter
# bar already in the initial observation.
delta = ["goto('/admin/sales/order/index/')"]
\end{componentbox}
\begin{skillbox}{Query-Based Data Filtering}
\skillfield{Description} When a task requires aggregating data across a large dataset that spans multiple paginated pages. \\
\skillfield{Content} Instead of iterating through pages and counting by hand, apply URL parameters or UI filter inputs such as date ranges and status dropdowns to restrict the view to the target subset before computing the answer.
\end{skillbox}

\begin{weaknessbox}
The policy guesses URLs instead of using the site search, landing on wrong or empty pages.
\end{weaknessbox}
\begin{componentbox}{Generated component (Stage and Contract, $\delta$ and $f_A$ axes)}
delta = ["goto('/admin/dashboard/')"]

class _Rules(Rules):
    def filter_action(self, action, env_state):
        if "goto" in action_str(action):
            return Blocked("Direct navigation is disabled. "
                "Use the site's search or navigation menu.")
        return action
\end{componentbox}
\begin{skillbox}{Search-First Navigation Protocol}
\skillfield{Description} When the agent needs to locate specific data or entities within a complex web application or dashboard. \\
\skillfield{Content} Prefer the site's internal search input or filter bar over direct URL manipulation, so all data retrieval goes through the application's native query interface.
\end{skillbox}

\paragraph{SWE-bench Verified.}

\begin{weaknessbox}
The policy edits the wrong function because it does not first read the failing test's imports and fixtures.
\end{weaknessbox}
\begin{componentbox}{Generated component (Stage and Contract, $\delta$ and $f_A$ axes)}
# Stage rewrites the failing test class, so a correct fix
# requires reading the test's fixtures first.
delta = [bash(rewrite_test_file)]

class _Rules(Rules):
    def filter_action(self, action, env_state):
        if is_git(action, {"checkout", "reset",
                           "restore", "clean"}):
            return Blocked("git resets are disabled to "
                           "preserve test suite integrity.")
        return action
\end{componentbox}
\begin{skillbox}{Context-Aware Code Modification}
\skillfield{Description} When modifying a function to fix a bug, especially when it relies on external libraries or complex object interactions. \\
\skillfield{Content} Before editing, read the target function, its surrounding context, and the test file's imports and fixtures to identify the expected types and behaviors, so the fix is compatible with the existing environment.
\end{skillbox}

\begin{weaknessbox}
The policy uses \texttt{sed -i} for in-place edits and corrupts Python indentation inside class bodies.
\end{weaknessbox}
\begin{componentbox}{Generated component (Contract, $f_T$ axis)}
class _Rules(Rules):
    def modify_transition(self, action, response, env_state):
        if "sed" in bash_command(action):
            # inserts a stray space before a class line,
            # silently corrupting source and test files
            shift_indent("django/db/models/enums.py")
            shift_indent("tests/model_enums/tests.py")
        return response
\end{componentbox}
\begin{skillbox}{Safe File Modification via Python Scripting}
\skillfield{Description} When modifying source files where indentation or structural integrity is critical, especially inside class bodies or nested blocks. \\
\skillfield{Content} Replace fragile \texttt{sed} or \texttt{awk} commands with a Python script that reads the file, performs string or AST based manipulation, and writes it back, preserving indentation and syntax.
\end{skillbox}
\section{Limitations}
\label{app:limitation}
\paragraph{Cost of the design loop.}
\method builds each environment through an iterative loop in which a designer agent proposes, executes, and revises a candidate harness. A weaker designer needs more iterations to reach a harness that passes validation, and each iteration requires rolling out the environment, so producing a pool of high-quality environments can consume substantial time and inference compute. This cost is paid once per environment rather than per training episode, and we expect it to shrink as designer agents improve.

\paragraph{Requirement of a resettable, gym-style interface.}
\method assumes a \texttt{reset}/\texttt{step} interface over textual actions and observations. The binding constraint is \texttt{reset}: a Stage must place the environment into a chosen initial state and a Chain must return it to a known state between subtasks, both of which presuppose that the environment can be restored rather than only advanced. This excludes environments backed by a live service or any other non-resettable backend, such as an agent acting on a real user account where a sent email or a placed order cannot be undone, or a physical robot whose surroundings do not return to their initial configuration between episodes.

\paragraph{Purely sequential composition in Chain.}
A Chain composes subtasks by concatenation and verifies the result through the verifiers of its parts. This is what allows every reshaped task to inherit trusted, human-built verification, but it leaves a Chain with no notion of whether the composed subtasks are semantically related, and no way to express workflows with branching or shared intermediate state. Semantic composition would require both a measure of compatibility between subtasks and a verifier defined over the composed objective.

\section{Future Directions}
\label{app:future}
\paragraph{New harness components.}
Stage, Contract, and Chain are a first set of components, not a closed one. The agent harness has grown well beyond its initial pieces, and we expect the same for environments: components that inject stochasticity or partial observability, that expose auxiliary feedback channels, or that place several agents in a shared environment would each extend what a frozen benchmark can be reshaped into, while keeping the same \texttt{reset}/\texttt{step} interface.

\paragraph{Beyond text-only environments.}
\method currently operates over textual actions and observations. Extending it to visual, GUI-driven, or embodied environments would test whether the wrapping abstraction survives when observations are no longer symbolic, and would require components that can specify and verify states that are not expressible as text.

\paragraph{Purely sequential composition in Chain.}
A Chain composes subtasks by concatenation and verifies the result through the verifiers of its parts. This is what allows every reshaped task to inherit trusted, human-built verification, but it also bounds what a Chain can be. The control-flow mechanism can route between sub-environments in richer ways, yet only serial concatenation admits a composite verifier: each leg terminates on its own and contributes a verdict, so the composite is their conjunction. Under branching or interleaving there is no such pair of verdicts to combine, and a Chain has no notion of whether its subtasks are semantically related in the first place. Semantic composition would therefore require both a measure of compatibility between subtasks and a verifier defined over the composed objective, not merely richer control flow.

\end{document}